\documentclass[oneside]{book}

\usepackage[preprint]{neurips_2020_arxivadapted}

\usepackage[utf8]{inputenc} %
\usepackage[T1]{fontenc}    %
\usepackage{hyperref}       %
\usepackage{url}            %
\usepackage{booktabs}       %
\usepackage{amsfonts}       %
\usepackage{nicefrac}       %
\usepackage{microtype}      %
\usepackage{nameref}
\usepackage[shortlabels]{enumitem}
\usepackage{textgreek}

\usepackage{float}
\usepackage{graphicx}
\usepackage{tocstyle}

\usepackage{color}
\usepackage{amsmath}
\usepackage{amssymb}
\usepackage{amsthm}
\usepackage{mathtools}
\usepackage{thmtools}
\usepackage[mathscr]{eucal}%
\usepackage{bm}%
\usepackage{bbm}
\usepackage{relsize}
\usepackage{graphics}
\usepackage{wrapfig}
\usepackage{multirow}
\usepackage{enumitem} 
\usepackage{varwidth}

\usepackage{array}

\usepackage{pgf}
\usepackage{xinttools}%
\usepackage{tikz}
\usetikzlibrary{arrows.meta}
\usepackage{textcomp}
\usetikzlibrary{calc,shapes,arrows,positioning,automata,trees}
\usepackage{placeins} %
\usepackage{tabularx}
\usepackage{graphicx}
\usepackage{subcaption} %
\usepackage{listings}
\usepackage{xargs,lipsum,caption,changepage,ifthen}

\usepackage{tikz} %

\usepackage{titlesec}
\titleformat{\chapter}[display]
  {\normalfont\bfseries}{}{0pt}{\Huge}

\usepackage{bigdelim}
\usepackage{colortbl} %
\definecolor{mColor1}{rgb}{0.95,0.95,0.95}
\newcolumntype{a}{>{\columncolor{mColor1}}c}
\newcolumntype{b}{>{\columncolor{mColor1}}l}

\usepackage[toc,page]{appendix}

\newtheorem{theorem}{Theorem}
\newtheorem{definition}{Definition}
\newtheorem*{theorem*}{Theorem}
\newtheorem*{definition*}{Definition}

\newcommand\Bk{\bm{k}}

\newcommand\Bp{\bm{p}}
\newcommand\Bq{\bm{q}}

\newcommand\Bu{\bm{u}}
\newcommand\Bv{\bm{v}}

\newcommand\Bx{\bm{x}}
\newcommand\By{\bm{y}}
\newcommand\Bz{\bm{z}}

\newcommand\BK{\bm{K}}

\newcommand\BQ{\bm{Q}}

\newcommand\BV{\bm{V}}
\newcommand\BW{\bm{W}}
\newcommand\BX{\bm{X}}
\newcommand\BY{\bm{Y}}
\newcommand\BZ{\bm{Z}}

\newcommand\Bxi{\bm{\xi}}

 \newcommand{\dR}{\mathbb{R}}

\newcommand{\rE}{\mathrm{E}}

\newcommand{\rS}{\mathrm{S}}

\newcommand\nn{\mathrm{new}}

\newcommand\att{\mathrm{att}}

\newcommand{\NRM}[1]{{{\left\| #1\right\|}}} %
\renewcommand{\leq}{\leqslant}

\newcommand{\soft}{\mathrm{softmax}}

\makeatletter
\newcommand{\dlmf}[1]{%
\cite[%
  \def\nextitem{\def\nextitem{, }}%
  \@for \el:=#1\do{\nextitem\href{http://dlmf.nist.gov/\el}{(\el)}}%
]{Olver:10}%
}
\makeatother

\usepackage{pdflscape} %

\newcolumntype{R}[1]{>{\raggedright\arraybackslash}p{#1}}
\newcolumntype{C}[1]{>{\centering\arraybackslash}p{#1}}
\newcolumntype{L}[1]{>{\raggedleft\arraybackslash}p{#1}}

\usepackage{bigdelim}
\usepackage{colortbl} %
\definecolor{mColor1}{rgb}{0.95,0.95,0.95}

\makeatletter
\newcommand{\chapterauthor}[1]{%
  {\parindent0pt\vspace*{-25pt}%
  \linespread{1.1} \large\bf#1%
  \par\nobreak\vspace*{35pt}}
  \@afterheading%
}
\makeatother

\hypersetup{
   colorlinks=true,
   linkcolor=blue,
   citecolor=green,
   urlcolor=magenta,
   pdfborder=0 0 0,
   pdftitle={Modern Hopfield Networks and Attention for Immune Repertoire Classification},
   pdfsubject={Modern Hopfield Networks, Convergence, 
               Storage Capacity, Antibody, Immune Repertoire}, 
   pdfkeywords={Hopfield Network, Attention,  Continuous State,
                Energy, Convergence, Storage Capacity, Antibody, Immune Repertoire, Machine Learning},
   pdfauthor={Michael Widrich, Bernhard Sch\"{a}fl, Hubert Ramsauer,            Milena Pavlovi{\'c},
              Lukas Gruber, Markus Holzleitner, Johannes Brandstetter, 
              Geir Kjetil Sandve, Victor Greiff, 
              Sepp Hochreiter, G\"{u}nter Klambauer 
              },%
   pdfstartview=FitH
}

\title{Modern Hopfield Networks and Attention for Immune Repertoire Classification}

\author{
    Michael Widrich\footnotemark[1] \And 
    Bernhard Sch\"{a}fl\footnotemark[1]  \And 
    Milena Pavlovi{\'c}\footnotemark[3]~$~^{,}$\footnotemark[4]\And 
    Hubert Ramsauer\footnotemark[1]  \And 
    Lukas Gruber\footnotemark[1]  \And  
    Markus Holzleitner\footnotemark[1]  \And  
    Johannes Brandstetter\footnotemark[1]  \And 
    Geir Kjetil Sandve\footnotemark[4]  \And 
    Victor Greiff\footnotemark[3] \And 
    Sepp Hochreiter\footnotemark[1]~$~^{,}$\footnotemark[2] \And 
    G\"{u}nter Klambauer\footnotemark[1]  \\
  \footnotemark[1]~~ELLIS Unit Linz and LIT AI Lab, \\ 
                    Institute for Machine Learning,\\
                    Johannes Kepler University Linz, Austria \\
  \footnotemark[2]~~Institute of Advanced Research in 
                    Artificial Intelligence (IARAI)\\
  \footnotemark[3]~~Department of Immunology, University of Oslo, Norway\\ 
  \footnotemark[4]~~Department of Informatics, University of Oslo, Norway
}

\DeclareUnicodeCharacter{2212}{-}

\begin{document}

\maketitle

\begin{abstract}
A central mechanism in machine learning is to identify, store, and recognize patterns.
How to learn, access, and retrieve such patterns is crucial in Hopfield networks and the more recent transformer architectures.
We show that the attention mechanism of transformer architectures is actually the update rule of
modern Hopfield networks that can store exponentially many patterns. 
We exploit this high storage capacity of modern Hopfield networks
to solve a challenging multiple instance learning (MIL)
problem in computational biology:
immune repertoire classification.  
Accurate and interpretable machine learning methods solving this problem
could pave the way towards new vaccines and therapies,
which is currently a very relevant
research topic intensified by the COVID-19 crisis.
Immune repertoire classification based on the vast number of 
immunosequences of an individual is a MIL problem with 
an unprecedentedly massive number of instances, two orders of magnitude larger than 
currently considered problems, and with an extremely low witness rate.
In this work, we present our novel method DeepRC that 
integrates transformer-like attention, or equivalently modern Hopfield networks,
into deep learning architectures for massive MIL such as immune repertoire classification. 
We demonstrate that DeepRC outperforms all other methods 
with respect to predictive performance on large-scale experiments,
including simulated and real-world virus infection data,
and enables the extraction of 
sequence motifs that are connected to a given disease class.
Source code and datasets: \textit{https://github.com/ml-jku/DeepRC}
\end{abstract}

\def\cheatspace{0mm} %

\vspace{\cheatspace}
\section*{Introduction}
\vspace{\cheatspace}
\label{introduction}
Transformer architectures \citep{vaswani2017attention} and their attention mechanisms are currently used
in many applications, such as natural language processing (NLP), imaging, and 
also in multiple instance learning (MIL) problems \citep{lee2019set}. 
In MIL, a set or bag of objects is labelled rather than objects 
themselves as in 
standard supervised learning tasks \citep{dietterich1997solving}. 
Examples for MIL problems are 
medical images, in which each sub-region of the image represents an instance,
video classification, in which each frame is an instance,
text classification, where words or sentences are instances of a text,
point sets, where each point is an instance of a 3D object, and
remote sensing data, where each sensor is an instance \citep{carbonneau2018survey,uriot2019learning}.
Attention-based MIL has been successfully used for
image data, for example to identify tiny objects 
in large images \citep{ilse2018attention,pawlowski2019needles,tomita2019attention,kimeswenger2019detecting}
and transformer-like attention mechanisms for sets of points and images \citep{lee2019set}. 

However, in MIL problems considered by machine learning methods
up to now, the number of instances per bag is in the range of hundreds or few thousands \citep{carbonneau2018survey,lee2019set} (see also Tab.~\ref{tab:mil_datasets}).
At the same time the witness rate (WR), the rate of discriminating instances per bag, 
is already considered low at $1\%-5\%$. We will tackle 
the problem of \emph{immune repertoire classification} with hundreds of thousands of instances 
per bag without instance-level labels and with extremely low witness rates
down to $0.01\%$ using an attention mechanism.
We show that the attention mechanism of transformers 
is the update rule of modern Hopfield networks \citep{Krotov:16,Krotov:18,Demircigil:17}
that are generalized to continuous states in contrast to classical Hopfield networks
\citep{Hopfield:82}.
A detailed derivation and analysis of
modern Hopfield networks is given 
in our companion paper \citep{Ramsauer:20}.
These novel continuous state Hopfield networks allow to store and retrieve
exponentially (in the dimension of the space) many patterns 
(see next Section). %
Thus, modern Hopfield networks with their update rule, which are 
used as an attention mechanism in the transformer, enable 
immune repertoire classification in computational biology.

\begin{figure}[t]
    \centering
    \includegraphics[width=0.8\textwidth]{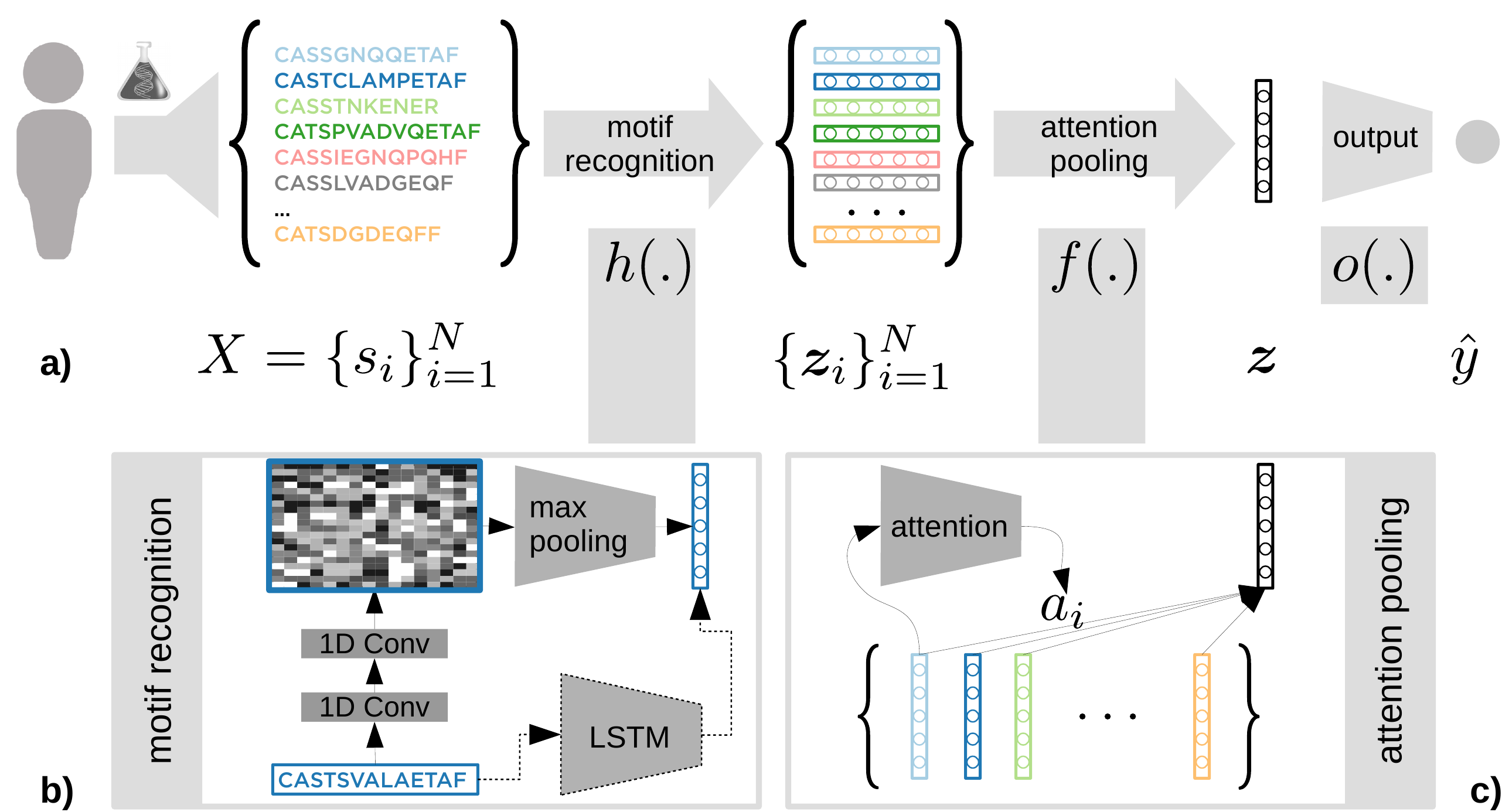}
    \caption[]{Schematic representation of the DeepRC approach.
    \textbf{a)} An immune repertoire $X$ is represented by large bags of immune receptor sequences (colored). 
    A neural network (NN) $h$ serves to recognize patterns in
    each of the sequences $s_i$ and maps them to sequence-representations $\Bz_i$.
    A pooling function $f$ is used to obtain a repertoire-representation $\Bz$ for the input
    object.
    Finally, an output network $o$ predicts the class label $\hat y$. 
    \textbf{b)} DeepRC uses stacked 1D convolutions for a parameterized function $h$ due to their computational efficiency. Potentially, millions 
    of sequences have to be processed for each input object. In principle, also 
    recurrent neural networks (RNNs), such as LSTMs \citep{hochreiter2007fast}, or transformer 
    networks \citep{vaswani2017attention} may be used 
    but are currently computationally too costly. 
    \textbf{c)} Attention-pooling is used to obtain a repertoire-representation $\Bz$ for 
    each input object, where DeepRC uses weighted averages of sequence-representations. 
    The weights are determined by an update rule of modern Hopfield networks that 
    allows to retrieve exponentially many patterns.
    } 
    \vspace{-5mm}
    \label{fig:deeprc_approach}
\end{figure}

Immune repertoire classification, i.e.\ classifying the immune status 
based on the immune repertoire sequences, is essentially a text-book example for  
a \emph{multiple instance learning} problem \citep{dietterich1997solving,maron1998framework,wang2018revisiting}.
Briefly, the immune repertoire of an individual
consists of an immensely large bag of 
immune receptors, represented as amino acid sequences.
Usually, the presence of only a small fraction 
of particular receptors determines the immune 
status with respect to a particular disease \citep{christophersen2014tetramer, emerson2017immunosequencing}.
This is because 
the immune system has already acquired a resistance 
if one or few particular immune receptors that can bind to the disease agent are present.
Therefore, classification of immune repertoires
bears a high difficulty since each immune repertoire can contain millions
of sequences as instances with only a few indicating the class.
Further properties of the data that complicate the problem are:
(a) The overlap of immune repertoires of different individuals is low (in most cases, maximally low single-digit percentage values) \citep{greiff2017learning, elhanati2018predicting},
(b) multiple different sequences can bind to the same pathogen \citep{wucherpfennig2007polyspecificity}, and 
(c) only subsequences within the sequences determine
whether binding to a pathogen is possible \citep{dash2017quantifiable,glanville2017identifying, akbar2019compact,springer2020tcrpeptidebinding,fischer2019predicting}. 
In summary, immune repertoire classification can be formulated 
as multiple instance learning with an extremely low witness rate and large numbers
of instances, which represents a challenge for currently available machine learning methods.
Furthermore, the methods should ideally be interpretable, since 
the extraction of class-associated sequence motifs is desired to gain 
crucial biological insights.

The acquisition of human immune repertoires has been enabled 
by immunosequencing technology \citep{georgiou2014promise, brown2019augmenting}
which allows to obtain the immune receptor sequences and immune repertoires
of individuals. Each individual is uniquely characterized by their immune repertoire, which
is acquired and changed during life. This repertoire may be influenced by
all diseases that an individual is exposed to during their lives and hence
contains highly valuable information about those diseases and the individual's immune status. 
Immune receptors enable the immune system to specifically recognize disease agents or pathogens. 
Each immune encounter is recorded as an immune event into immune memory by preserving and
amplifying immune receptors in the repertoire used to fight a given disease. 
This is, for example, the working principle of vaccination.
Each human has about $10^7$--$10^8$ unique immune receptors with low overlap
across individuals and sampled from a potential diversity of $>10^{14}$ 
receptors \citep{mora2019howmanyifferentclonotypes}. 
The ability to sequence and analyze human immune receptors at large scale 
has led to fundamental and mechanistic insights into the adaptive immune 
system and has also opened the opportunity for the development of novel 
diagnostics and therapy approaches \citep{georgiou2014promise, brown2019augmenting}.

Immunosequencing data have been analyzed with computational 
methods for a variety
of different tasks \citep{greiff2015bioinformatic,shugay2015vdjtools,miho2018computational,yaari2015practical, wardemann2017novel}. 
A large part of the available machine learning methods for immune 
receptor data has been focusing on the individual 
immune receptors in a repertoire, with the aim to, for example, predict the antigen 
or antigen portion (epitope) to which these sequences
bind or to predict sharing of receptors across individuals \citep{gielis2019tcrex,springer2020tcrpeptidebinding,jurtz2018nettcr,moris2019treating, fischer2019predicting, greiff2017learning, sidhom2019deeptcr,elhanati2018predicting}.
Recently, \citet{jurtz2018nettcr} used 1D convolutional 
neural networks (CNNs) to predict antigen binding of T-cell receptor (TCR) sequences 
(specifically, binding of TCR sequences to peptide-MHC complexes) 
and demonstrated that motifs can be extracted
from these models. 
Similarly, \citet{konishi2019capturing} use CNNs, gradient boosting, 
and other machine learning techniques on B-cell receptor (BCR) sequences to distinguish
tumor tissue from normal tissue.
However, the methods presented so far predict a particular class, 
the epitope, based on a single input sequence.

Immune repertoire classification has been considered
as a MIL problem in the following publications.
A Deep Learning framework called DeepTCR \citep{sidhom2019deeptcr}
implements several Deep Learning approaches for immunosequencing data. 
The computational framework, inter alia,
allows for attention-based MIL repertoire classifiers 
and implements a basic form of attention-based averaging.
\citet{ostmeyer2019biophysicochemical} already suggested a MIL method for 
immune repertoire classification. This method considers 4-mers, 
fixed sub-sequences 
of length 4, as instances of an input object and trained a logistic 
regression model with these 4-mers as input. 
The predictions of the logistic regression model for each 4-mer were max-pooled to
obtain one prediction per input object.
This approach is characterized by  
(a) the rigidity of the k-mer features as compared to convolutional 
kernels \citep{alipanahi2015predicting, zhou2015predicting, zeng2016convolutional}, 
(b) the max-pooling operation, which constrains the network to learn from a 
single, top-ranked k-mer for each iteration over the input object, and
(c) the pooling of prediction scores rather than representations \citep{wang2018revisiting}.
Our experiments also support that these 
choices in the design of the method can lead to constraints on the 
predictive performance (see Table~\ref{tab:results_full}). 

Our proposed method, DeepRC, also uses a MIL approach
but considers sequences rather than k-mers as instances within an input object
and a transformer-like attention mechanism.
DeepRC sets out to avoid the above-mentioned constraints of current methods by
(a) applying transformer-like attention-pooling instead of max-pooling and 
learning a classifier on the repertoire rather 
than on the sequence-representation,
(b) pooling learned representations rather than predictions, and
(c) using less rigid feature extractors,
such as 1D convolutions or LSTMs.
\emph{In this work, we contribute the following:}
We demonstrate that continuous generalizations of binary 
modern Hopfield-networks \citep{Krotov:16, Krotov:18, Demircigil:17} 
have an update rule that is known as the attention mechanisms
in the transformer. We show that these modern Hopfield networks have exponential storage
capacity, which allows them to extract patterns among a large set of instances (next Section). %
Based on this result, 
we propose DeepRC, 
a novel deep MIL method based on 
modern Hopfield networks for large bags
of complex sequences,
as they occur in immune repertoire classification (Section "Deep Repertoire Classification). %
We evaluate the predictive performance of DeepRC and other machine learning approaches for the 
classification of immune repertoires in a large comparative study (Section "Experimental Results")%

\vspace{\cheatspace}
\section*{Exponential storage capacity of continuous state modern Hopfield networks with
transformer attention as update rule}
\label{sec:theory}
\vspace{\cheatspace}
In this section,
we show that modern Hopfield networks have exponential storage 
capacity, which will later allow us to approach 
massive multiple-instance learning problems, such as immune repertoire classification.
See our companion paper \citep{Ramsauer:20} for a detailed derivation and analysis of
modern Hopfield networks. 
We assume patterns $\Bx_1,\ldots,\Bx_N \in \dR^d$
that are stacked as columns to 
the matrix $\BX \ = \ \left( \Bx_1,\ldots,\Bx_N \right)$ and a 
query pattern $\Bxi$ that also represents the current state. 
The largest norm of a pattern is
$M \ = \ \max_{i} \NRM{\Bx_i}$. The \emph{separation} $\Delta_i$  of a 
pattern $\Bx_i$ is defined as its minimal dot product difference to any of the other 
patterns:
$\Delta_i = \min_{j,j \not= i} \left( \Bx_i^T \Bx_i - \Bx_i^T \Bx_j \right)$. 
A pattern is \emph{well-separated} from the data if $
 \Delta_i  \geq \frac{2}{\beta N} + \frac{1}{\beta} \log \left( 2 (N-1)  N  \beta  M^2 \right)$.
We consider a modern Hopfield network with current state $\Bxi$
and the energy function 
$\rE  =   -  \beta^{-1} \log \left( \sum_{i=1}^N
\exp(\beta \Bx_i^T \Bxi) \right)   +  \beta^{-1} \log N  + 
\frac{1}{2} \Bxi^T \Bxi  +  \frac{1}{2} M^2$. 
For energy $\rE$ and state $\Bxi$, the update rule
\begin{align}
\label{eq:main_iterate}
\Bxi^{\nn} \ &= \ f(\Bxi;\BX,\beta) \ = \ \BX \ \Bp \ = \   \BX \ \soft ( \beta \BX^T \Bxi)
\end{align}
is proven to converge globally  
to stationary points of the energy $\rE$, which are local minima or saddle points
(see \citep{Ramsauer:20}, appendix, Theorem A2 ). %
{\em Surprisingly, the update rule Eq.~\eqref{eq:main_iterate}
is also the formula of the well-known transformer attention mechanism.}

To see this more clearly, we simultaneously update several queries $\Bxi_i$.
Furthermore the queries $\Bxi_i$ and the patterns $\Bx_i$ are 
linear mappings of vectors $\By_i$ into the space $\dR^d$.
For matrix notation,
we set $\Bx_i = \BW_K^T \By_i$, $\Bxi_i = \BW_Q^T \By_i$
and multiply the result of our update rule with $\BW_V$.
Using $\BY=(\By_1,\ldots,\By_N)^T$, we define
the matrices $\BX^T=\BK = \BY \BW_K $, $\BQ = \BY \BW_Q$,
and $\BV=\BY \BW_K \BW_V=\BX^T \BW_V$, where 
$\BW_K \in \dR^{d_y\times d_k}, \BW_Q \in \dR^{d_y\times d_k}, \BW_V \in \dR^{d_k\times d_v}$,
$\BK \in \dR^{N \times d_k}$, $\BQ \in \dR^{N \times d_k}$, $\BV \in \dR^{N \times d_v}$, and
the patterns are now mapped to the Hopfield space with dimension $d=d_k$. 
We set $\beta = 1/\sqrt{d_k}$ and change
$\soft$ to a row vector.
The update rule Eq.~\eqref{eq:main_iterate} multiplied by $\BW_V$
performed for all queries simultaneously becomes in row vector notation:
\begin{align}
 \label{eq:attention}
  \att(\BQ,\BK,\BV; \beta) \ &= \ \soft \left( \beta \ \BQ \ \BK^T \right) \ \BV \ = \ \soft \left( \left( 1/\sqrt{d_k} \right) \ \BQ \ \BK^T \right) \ \BV \ .
\end{align}
This formula is the transformer attention.
If the patterns $\Bx_i$ are well separated, the iterate Eq.~\eqref{eq:main_iterate}
converges to a fixed point close to a pattern to which the initial $\Bxi$ is similar. 
If the patterns are not well separated 
the iterate Eq.\eqref{eq:main_iterate} converges to 
a fixed point close to the arithmetic mean of the patterns. 
If some patterns are 
similar to each other but well separated from all
other vectors, then a {\em metastable state} between the similar
patterns exists. Iterates that start near a metastable state converge
to this metastable state.
For details see \cite{Ramsauer:20}, appendix, Sect.~A2. %
Typically, the update converges after one update step
(see \cite{Ramsauer:20}, appendix, Theorem~A8) %
and has an exponentially small retrieval error 
(see \cite{Ramsauer:20}, appendix, Theorem~A9). %
Our main concern for application to 
immune repertoire classification 
is the number of patterns that can be 
stored and retrieved by the modern Hopfield network, equivalently to 
the transformer attention head. 
The storage capacity of an attention mechanism is critical for 
massive MIL problems.
We first define what we mean by storing and retrieving patterns
from the modern Hopfield network.
\begin{definition}[Pattern Stored and Retrieved]
We assume that around every pattern $\Bx_i$ a sphere $\rS_i$ is given.
We say {\em $\Bx_i$ is stored} if there is 
a single fixed point $\Bx_i^* \in \rS_i$ to
which all points $\Bxi \in \rS_i$ converge,
and  $\rS_i \cap \rS_j = \emptyset$ for $i \not= j$.
We say {\em $\Bx_i$ is retrieved} if 
the iteration Eq.~\eqref{eq:main_iterate}
converged to the single fixed point $\Bx_i^* \in \rS_i$.
\end{definition}

For randomly chosen patterns, the number of patterns that can be stored
is exponential in the dimension $d$ of the space of the patterns ($\Bx_i \in \dR^d$).
\begin{theorem}
\label{th:2}
We assume a failure probability $0<p\leq 1$ and randomly chosen patterns 
on the sphere with radius $M=K \sqrt{d-1}$. 
We define  $ a \ := \  \frac{2}{d-1} \ (1 \ + \ \ln(2 \ \beta \ K^2 \  p \  (d-1)))$, 
$b \ := \ \frac{2 \ K^2 \ \beta}{5}$,
and $c \ = \ \frac{b}{W_0(\exp(a \ + \ \ln(b))}$,
where $W_0$ is the upper branch of the Lambert $W$ function 
and ensure $c \ \geq \ \left( \frac{2}{ \sqrt{p}}\right)^{\frac{4}{d-1}}$.
Then with probability $1-p$, the number of random patterns 
that can be stored is  
\begin{align} 
    N \ &\geq \ \sqrt{p} \ c^{\frac{d-1}{4}}  \ .
\end{align}

Examples are $c\geq 3.1546$ for
$\beta=1$, $K=3$, $d= 20$ and $p=0.001$ ($a + \ln(b)>1.27$)
and  $c\geq 1.3718$ for $\beta = 1$ $K=1$, $d = 75$, and $p=0.001$
($a + \ln(b)<-0.94$).
\end{theorem}
See \cite{Ramsauer:20}, appendix, Theorem~A5 %
for a proof. We have established that
a modern Hopfield network or a transformer attention mechanism can store
and retrieve exponentially many patterns.
This allows us to approach
MIL with massive numbers of instances from
which we have to retrieve a few with an attention mechanism.

\vspace{\cheatspace}
\section*{Deep Repertoire Classification} 
\vspace{\cheatspace}
\label{sec:deeprc}
\textbf{Problem setting and notation.} We consider a MIL problem, in 
which an input object $X$ is a \emph{bag} of 
$N$ instances $X=\{s_1,\ldots,s_{N}\}$. The instances 
do not have dependencies nor orderings between them
and $N$ can be different for every object. 
We assume that each instance $s_{i}$ is associated with 
a label $y_i \in \{0,1\}$, assuming a binary classification task, 
to which we do not have access. We only
have access to a label $Y=\max_i{y_i}$ for an input object
or bag.
Note that this poses a credit assignment problem, since the
sequences that are responsible for the label $Y$ have to be identified
and that the relation between instance-label and bag-label can be more
complex \citep{foulds2010review}.

A model $\hat y = g(X)$ should be 
(a) invariant to permutations of the instances and 
(b) able to cope with the fact that $N$ varies across input objects \citep{ilse2018attention}, which
is a problem also posed by point sets \citep{qi2017pointnet}. 
Two principled approaches exist. The first approach is to learn an instance-level
scoring function $h:\mathcal S \mapsto [0,1]$, which is then pooled 
across instances with a pooling function $f$, 
for example by average-pooling or max-pooling (see below).
The second approach is to construct an instance representation $\Bz_i$ of each instance by 
$h:\mathcal S \mapsto \mathbb R^{d_v}$ and then 
encode the bag, or the input object, 
by pooling these instance representations \citep{wang2018revisiting}
via a function $f$. 
An output function $o:\mathbb R^{d_v} \mapsto [0,1]$ subsequently classifies the bag. 
The second approach, the pooling of representations rather than scoring functions, is currently best performing \citep{wang2018revisiting}.

In the problem at hand, the input object $X$ is the 
immune repertoire of an individual that consists of a large set of 
immune receptor sequences (T-cell receptors or antibodies). 
Immune receptors are primarily 
represented as sequences $s_i$ from a space $s_i \in \mathcal S$.
These sequences act as the instances in the MIL problem. 
Although immune repertoire classification can readily be 
formulated as a MIL problem,
it is yet unclear how well machine learning 
methods solve the above-described problem with 
a large number of instances $N \gg 10,000$ and with 
instances $s_i$ being complex sequences.
Next we describe currently used
pooling functions for MIL problems.

\textbf{Pooling functions for MIL problems.} 
Different pooling functions equip a model $g$ with the property 
to be invariant to permutations of instances and with the ability 
to process different numbers of instances. 
Typically, a neural network $h_{\bm \theta}$ 
with parameters $\bm \theta$ is trained 
to obtain a function that maps each instance onto a representation:
$\bm z_i = h_{\bm \theta}(s_i)$ and then a pooling function
$\Bz=f(\{\Bz_1,\ldots,\Bz_N\})$ supplies a 
representation $\Bz$ of the input object $X=\{s_1,\ldots,s_{N}\}$.
The following pooling functions are typically used:
\emph{average-pooling}: $\bm z=\frac{1}{N}\sum_{i=1}^{N} \bm z_i$,
\emph{max-pooling}: $\bm z= \sum_{m=1}^{d_v} \bm e_m (\max_{i, 1\leq i \leq N} \{ z_{im} \}),$
    where $\bm e_m$ is the standard basis vector for dimension $m$ 
    and
\emph{attention-pooling}: $\bm z=\sum_{i=1}^{N} a_i \bm z_i$,
where $a_i$ are non-negative ($a_i\geq0$), sum to one  ($\sum_{i=1}^{N} a_i=1$), and are determined by an attention mechanism.
These pooling functions are invariant to permutations of $\{1,\dots,N\}$ 
and are differentiable.
Therefore, they are suited as building blocks 
for Deep Learning architectures. 
We employ attention-pooling in our DeepRC model as detailed
in the following.

\textbf{Modern Hopfield networks viewed as transformer-like attention mechanisms.}
The modern Hopfield networks, as introduced above,%
have a storage capacity that is exponential in the dimension
of the vector space and converge after just
one update (see \citep{Ramsauer:20}, appendix).%
Additionally, the update rule of modern Hopfield networks is 
known as key-value attention mechanism,
which has been highly successful through the transformer \citep{vaswani2017attention} and BERT
\citep{Devlin:19} models in natural language 
processing.
Therefore using modern Hopfield networks with
the key-value-attention mechanism as update rule is 
the natural choice for our task.
In particular, modern Hopfield networks are 
theoretically justified for storing and retrieving
the large number of vectors (sequence patterns) that appear 
in the immune repertoire classification task. 

Instead of using the terminology of modern Hopfield networks,
we explain our DeepRC architecture in terms of 
key-value-attention
(the update rule of the modern Hopfield network),
since it is well known in the deep learning community.
The attention mechanism assumes a space of dimension $d_k$
in which keys and queries are compared.
A set of $N$ key vectors  
are combined to the matrix 
$\BK$.
A set of $d_{q}$ query vectors
are combined to the matrix $\BQ$.
Similarities between queries and keys are computed by inner products,
therefore queries can search for similar keys that are stored.
Another set of $N$ value vectors  
are combined to the matrix $\BV$.
The output of the attention mechanism is 
a weighted average of the value vectors for each query $\Bq$.
The $i$-th vector $\Bv_i$ is weighted by the similarity 
between the $i$-th key $\Bk_i$ and the query $\Bq$.
The similarity is given by the softmax of the inner products of 
the query $\Bq$ with the keys $\Bk_i$. 
All queries are calculated in parallel via matrix operations.
Consequently, the attention function $\mathrm{att}(\BQ,\BK,\BV;\beta)$ 
maps queries $\BQ$, keys $\BK$, and values $\BV$ to 
$d_{v}$-dimensional outputs: $\att(\BQ,\BK,\BV; \beta) = \soft(\beta \BQ \BK^T) \BV$ (see also Eq.~\eqref{eq:attention}). 
While this attention mechanism has originally been developed for
sequence tasks \citep{vaswani2017attention}, it can be readily 
transferred to sets \citep{lee2019set,ye2018learning}.
This type of attention mechanism will be employed in DeepRC.

\begin{wrapfigure}[36]{r}{0.45\textwidth}
  \begin{center}
    \includegraphics[width=0.45\textwidth]{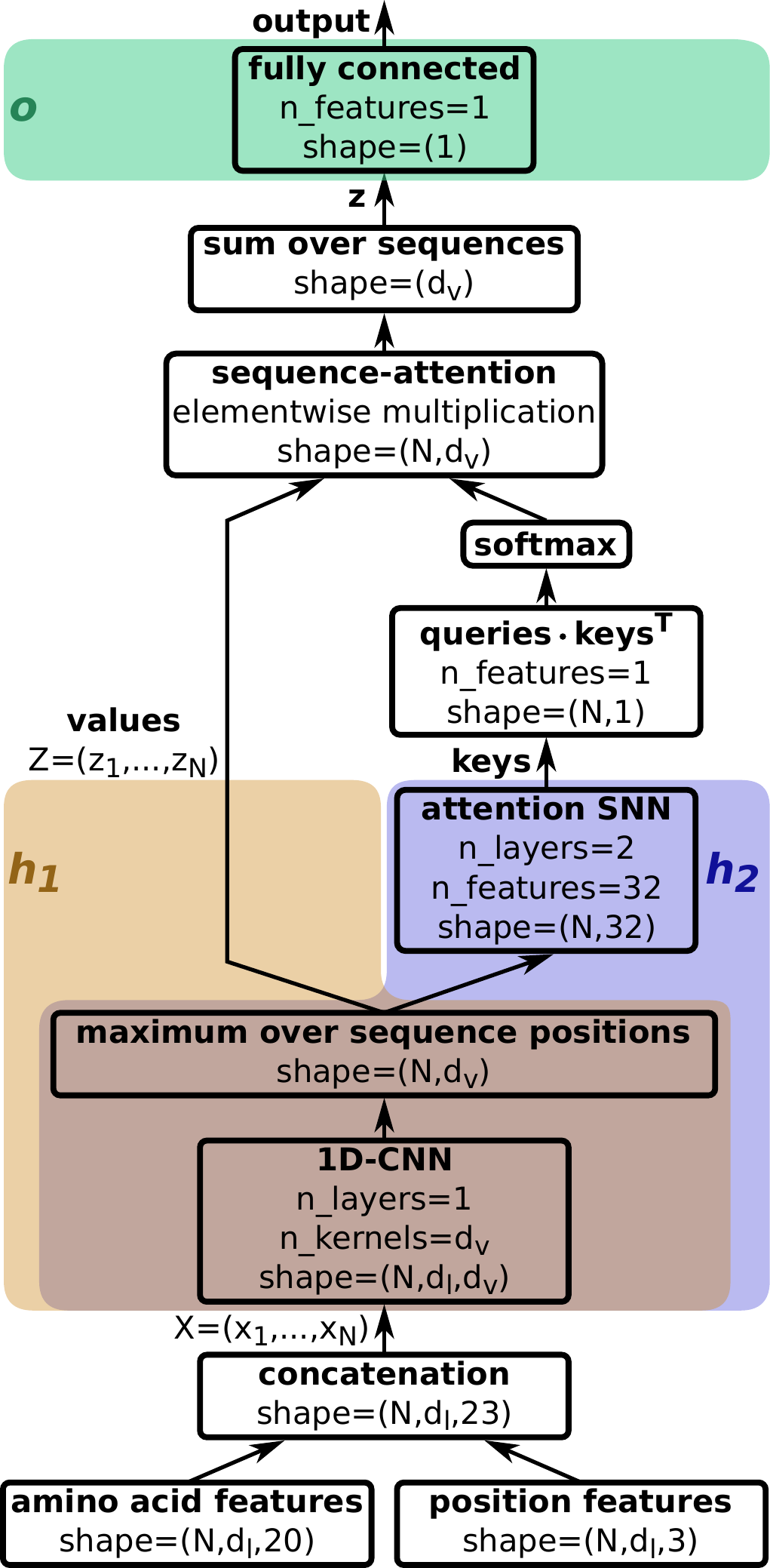}
  \end{center}
  \caption[]{DeepRC architecture as used in Table~\ref{tab:results_full} with sub-networks $h_1$, $h_2$, and $o$. $d_l$ indicates the sequence length.
   \label{fig:deeprc_architecture}}
\end{wrapfigure}

\textbf{The DeepRC method.} We propose a novel 
method \textbf{Deep} \textbf{R}epertoire \textbf{C}lassification 
(\textbf{DeepRC}) for immune 
repertoire classification with attention-based deep massive 
multiple instance learning and compare it against 
other machine learning approaches. 
For DeepRC, we consider \emph{immune repertoires as input objects},
which are represented as bags of instances. In a bag, \emph{each instance is
an immune receptor sequence} and each bag can contain a large number of sequences. 
Note that we will use $\Bz_i$ to denote the \emph{sequence-representation} of 
the $i$-th sequence and $\Bz$ to denote the \emph{repertoire-representation}. 
At the core, DeepRC consists of a transformer-like 
attention mechanism that
extracts the most important information from each repertoire.
We first give an overview of the attention mechanism and then 
provide details on each of the sub-networks 
$h_1$, $h_2$, and $o$ of DeepRC.
(Overview: Fig.~\ref{fig:deeprc_approach}; 
Architecture: Fig.~\ref{fig:deeprc_architecture}; 
Implementation details: Sect.~\ref{sec:deeprc_implementation_details};
DeepRC variations: Sect.~\ref{sec:deeprc_variations}.)

\textbf{Attention mechanism in DeepRC.}
This mechanism is based on the three
matrices $\BK$ (the keys), $\BQ$ (the queries), and $\BV$
(the values) together with a parameter
$\beta$.
\textit{Values.} 
DeepRC uses a 1D convolutional network $h_1$ \citep{lecun1998gradient,hu2014convolutional,kelley2016basset}
that supplies a sequence-representation $\Bz_i=h_1(s_i)$,
which acts as the values $\BV=\BZ= (\Bz_1, \ldots, \Bz_{N})$ 
in the attention mechanism (see Figure~\ref{fig:deeprc_architecture}). 
\textit{Keys.}
A second neural network $h_2$,
which shares its first layers with $h_1$,
is used to obtain keys 
$\BK \in \dR^{N \times d_k}$ for each 
sequence in the repertoire.
This network uses 2 self-normalizing layers \citep{klambauer2017self} 
with $32$ units per layer (see Figure~\ref{fig:deeprc_architecture}).
\textit{Query.} 
We use a fixed $d_k$-dimensional query vector $\Bxi$ which 
is learned via backpropagation.
For more attention heads, each head has a fixed query vector.
With the quantities introduced above, 
the transformer attention mechanism (Eq.~\eqref{eq:attention}) 
of DeepRC is implemented 
as follows:
\begin{align}
    \Bz= \att(\Bxi^T,\BK,\BZ; \frac{1}{\sqrt{d_k}}) = \soft\left(\frac{\Bxi^T \BK^T}{\sqrt{d_k}}\right) \BZ,
\end{align}
where $\BZ \in \dR^{N \times d_v}$ are the sequence--representations
stacked row-wise, $\BK$ are the keys, and
$\Bz$ is the repertoire-representation
and at the same time a weighted mean of sequence--representations $\Bz_i$. 
The attention mechanism can readily be extended to multiple queries, however,
computational demand could constrain this depending
on the application and dataset. Theorem
\ref{th:2} demonstrates that this mechanism is able to retrieve a single 
pattern out of several hundreds of thousands.

\textit{Attention-pooling and interpretability.}
Each input object, i.e. repertoire, consists of a large 
number $N$ of sequences,
which are reduced to a single fixed-size feature vector of length $d_v$ 
representing the whole input object by an attention-pooling function. 
To this end, a transformer-like attention 
mechanism adapted to sets is realized in DeepRC which supplies $a_i$ -- the importance of 
the sequence $s_i$. This importance value is an interpretable quantity, 
which is highly desired for the immunological problem at hand. 
Thus, DeepRC allows for two forms of interpretability methods.
(a) A trained DeepRC model can compute attention weights $a_i$,
which directly indicate the importance of a sequence.
(b) DeepRC furthermore allows for the usage of contribution analysis 
methods, such as Integrated Gradients (IG) 
\citep{sundararajan2017axiomatic} or Layer-Wise Relevance 
Propagation \citep{montavon2018methods,arras2019explaining}. See Sect.~\ref{appsec:integrated_gradients} for details.

\textbf{Classification layer and network parameters.} 
The repertoire-representation $\Bz$ is then 
used as input for a fully-connected output network 
$\hat y= o(\Bz)$ that predicts the immune status, 
where we found it sufficient to train single-layer networks.
In the simplest case, DeepRC predicts a single target, the class label $y$,
e.g. the immune status of an immune repertoire,
using one output value.
However, since DeepRC is an end-to-end deep learning model, multiple 
targets may be predicted simultaneously in classification or 
regression settings or a mix of both.
This allows for the introduction of additional information into the 
system via auxiliary targets such as age, sex, or other metadata.

\textbf{Network parameters, training, and inference.}
DeepRC is trained using standard gradient descent methods 
to minimize a cross-entropy loss. 
The network parameters are $\bm \theta_1, \bm \theta_2, \bm \theta_o$ for the 
sub-networks $h_1, h_2$, and $o$, respectively, 
and additionally $\Bxi$. In more detail, we train DeepRC using 
Adam \citep{kingma2014adam} with a
batch size of $4$ and dropout of input sequences.

\textbf{Implementation.}
To reduce computational time,
the attention network first computes the attention weights $a_i$ for each sequence $s_i$ in a repertoire.
Subsequently, the top $10\%$ of sequences 
with the highest $a_i$ per repertoire 
are used to compute the weight updates and prediction.
Furthermore, computation of $\Bz_i$ is performed in 16-bit,
others in 32-bit precision to ensure numerical stability in the softmax. See Sect.~\ref{sec:deeprc_implementation_details} for details.

\vspace{\cheatspace}
\section*{Experimental Results}
\vspace{\cheatspace}
\label{sec:results}
In this section, we report and analyze the predictive power of DeepRC and the 
compared methods on several immunosequencing datasets. %
The ROC-AUC is used as the main metric for the predictive 
power.

\textbf{Methods compared.} We compared previous methods
for immune repertoire classification, \citep{ostmeyer2019biophysicochemical} 
(``Log. MIL (KMER)'', ``Log. MIL (TCRB)'') and a burden 
test \citep{emerson2017immunosequencing}, 
as well as the baseline methods Logistic Regression (``Log. Regr.''), 
k-nearest neighbour (``KNN''), and Support Vector Machines (``SVM'') 
with kernels designed for sets, such as the Jaccard kernel (``J'')
and the MinMax (``MM'') kernel \citep{ralaivola2005graph}. For the simulated
data, we also added baseline methods that search for the implanted
motif either in binary or continuous fashion (``Known motif b.'', ``Known motif c.'') 
assuming that this motif was known (for details, see Sect.~\ref{subsec:methods_compared}).

\textbf{Datasets.}
We aimed at constructing 
immune repertoire classification 
scenarios with varying degree of difficulties and 
realism in order to compare and analyze
the suggested machine learning methods.
To this end, we either use
simulated or experimentally-observed immune receptor sequences and we implant signals,
specifically, sequence motifs or sets thereof \citep{akbar2019compact, weber2019immunesim}, at different 
frequencies into sequences of repertoires of the positive class. 
These frequencies represent the \emph{witness rates} and range from
$0.01\%$ to $10\%$. Overall, we compiled four categories of datasets:
(a) simulated immunosequencing data with implanted signals,
(b) LSTM-generated immunosequencing data with implanted signals,
(c) real-world immunosequencing data with implanted signals, and 
(d) real-world immunosequencing data with known immune status, the \emph{CMV dataset} \citep{emerson2017immunosequencing}.
The average number of instances per bag, which is the number of sequences per immune repertoire,
is $\approx$300,000 except for category (c), in which 
we consider the scenario of low-coverage data with only 10,000
sequences per repertoire. The number of repertoires 
per dataset ranges from 785 to 5,000. In total, all datasets 
comprise $\approx$30 billion sequences or instances. This represents the 
largest comparative study on immune repertoire classification (see Sect.~\ref{sec:datasets}).

\textbf{Hyperparameter selection.}
We used a nested 5-fold cross validation (CV) procedure to estimate
the performance of each of the methods.
All methods could adjust 
their most important hyperparameters 
on a validation set in the inner loop of the procedure.
See Sect.~\ref{sec:hyperparams} for details.

\begingroup
\setlength{\tabcolsep}{2pt} %
\renewcommand{\arraystretch}{1.5} %

\begin{table}[htp]%
    \begin{center}%
        \resizebox{\textwidth}{!}{%
        \begin{tabular}{lcacacacacac}
         \toprule
        & Real-world & \multicolumn{4}{c}{ Real-world data with  implanted signals} & \multicolumn{5}{c}{ LSTM-generated data} &  Simulated\\
    & CMV & OM 1\% & OM 0.1\% & MM 1\% & MM 0.1\% & 10\% & 1\% & 0.5\% & 0.1\% & 0.05\%  & avg.\\
        \cmidrule(r){2-2}
        \cmidrule(r){3-6}
        \cmidrule(r){7-11}
        \cmidrule(r){12-12}
        DeepRC & {\bf 0.831} \footnotesize{$\pm$ 0.022} & {\bf 1.00} \footnotesize{$\pm$ 0.00} & {\bf 0.98}\footnotesize{$\pm$ 0.01} & {\bf 1.00}\footnotesize{$\pm$ 0.00} & {\bf 0.94}\footnotesize{$\pm$0.01} & {\bf 1.00}\footnotesize{$\pm$ 0.00} & {\bf 1.00}\footnotesize{$\pm$ 0.00} & {\bf 1.00}\footnotesize{$\pm$ 0.00} & {\bf 1.00}\footnotesize{$\pm$ 0.00} & {\bf 1.00}\footnotesize{$\pm$ 0.00} & {\bf 0.846}\footnotesize{$\pm$ 0.223} \\
        SVM (MM) & 0.825 \footnotesize{$\pm$ 0.022} & {\bf 1.00} \footnotesize{$\pm$ 0.00} & 0.58\footnotesize{$\pm$ 0.02} & {\bf 1.00}\footnotesize{$\pm$ 0.00} & 0.53\footnotesize{$\pm$0.02} & {\bf 1.00}\footnotesize{$\pm$ 0.00} & {\bf 1.00}\footnotesize{$\pm$ 0.00} &{\bf 1.00}\footnotesize{$\pm$ 0.00} & {\bf 1.00}\footnotesize{$\pm$ 0.00} & 0.99\footnotesize{$\pm$ 0.01} & 0.827\footnotesize{$\pm$ 0.210} \\
        SVM (J) & 0.546 \footnotesize{$\pm$ 0.021} & 0.99 \footnotesize{$\pm$ 0.00} & 0.53\footnotesize{$\pm$ 0.02} & {\bf 1.00}\footnotesize{$\pm$ 0.00} & 0.57\footnotesize{$\pm$0.02} & 0.98\footnotesize{$\pm$ 0.04} & {\bf 1.00}\footnotesize{$\pm$ 0.00} & {\bf 1.00}\footnotesize{$\pm$ 0.00} & 0.90\footnotesize{$\pm$ 0.04} & 0.77\footnotesize{$\pm$ 0.07} & 0.550\footnotesize{$\pm$ 0.080} \\
        KNN (MM) & 0.679 \footnotesize{$\pm$ 0.076} & 0.74 \footnotesize{$\pm$ 0.24} & 0.49\footnotesize{$\pm$ 0.03} & 0.67\footnotesize{$\pm$ 0.18} & 0.50\footnotesize{$\pm$0.02} & 0.70\footnotesize{$\pm$ 0.27} & 0.72\footnotesize{$\pm$ 0.26} & 0.73\footnotesize{$\pm$ 0.26} & 0.54\footnotesize{$\pm$ 0.16} & 0.52\footnotesize{$\pm$ 0.15} & 0.634\footnotesize{$\pm$ 0.129} \\
        KNN (J) & 0.534 \footnotesize{$\pm$ 0.039} & 0.65 \footnotesize{$\pm$ 0.16} & 0.48\footnotesize{$\pm$ 0.03} & 0.70\footnotesize{$\pm$ 0.20} & 0.51\footnotesize{$\pm$0.03} & 0.70\footnotesize{$\pm$ 0.29} & 0.61\footnotesize{$\pm$ 0.24} & 0.52\footnotesize{$\pm$ 0.16} & 0.55\footnotesize{$\pm$ 0.19} & 0.54\footnotesize{$\pm$ 0.19} & 0.501\footnotesize{$\pm$ 0.007} \\
        Log. regr. & 0.607 \footnotesize{$\pm$ 0.058} & {\bf 1.00} \footnotesize{$\pm$ 0.00} & 0.54\footnotesize{$\pm$ 0.04} & 0.99\footnotesize{$\pm$ 0.00} & 0.51\footnotesize{$\pm$0.04} & {\bf 1.00}\footnotesize{$\pm$ 0.00} & {\bf 1.00}\footnotesize{$\pm$ 0.00} & 0.93\footnotesize{$\pm$ 0.15} & 0.60\footnotesize{$\pm$ 0.19} & 0.43\footnotesize{$\pm$ 0.16} & 0.826\footnotesize{$\pm$ 0.211} \\
        Burden test & 0.699 \footnotesize{$\pm$ 0.041} & {\bf 1.00} \footnotesize{$\pm$ 0.00} & 0.64\footnotesize{$\pm$ 0.05} & {\bf 1.00}\footnotesize{$\pm$ 0.00} & 0.89\footnotesize{$\pm$0.02} & {\bf 1.00}\footnotesize{$\pm$ 0.00} & {\bf 1.00}\footnotesize{$\pm$ 0.00} & {\bf 1.00}\footnotesize{$\pm$ 0.00} & {\bf 1.00}\footnotesize{$\pm$ 0.00} & 0.79\footnotesize{$\pm$ 0.28} & 0.549\footnotesize{$\pm$ 0.074} \\
        Log. MIL (KMER) & 0.582 \footnotesize{$\pm$ 0.065} & 0.54 \footnotesize{$\pm$ 0.07} & 0.51\footnotesize{$\pm$ 0.03} & 0.99\footnotesize{$\pm$ 0.00} & 0.62\footnotesize{$\pm$0.15} & {\bf 1.00}\footnotesize{$\pm$ 0.00} & 0.72\footnotesize{$\pm$ 0.11} & 0.64\footnotesize{$\pm$ 0.14} & 0.57\footnotesize{$\pm$ 0.15} & 0.53\footnotesize{$\pm$ 0.13} & 0.665\footnotesize{$\pm$ 0.224} \\
        Log. MIL (TCR\textbeta) & 0.515 \footnotesize{$\pm$ 0.073} & 0.50 \footnotesize{$\pm$ 0.03} & 0.50\footnotesize{$\pm$ 0.02} & 0.99\footnotesize{$\pm$ 0.00} & 0.78\footnotesize{$\pm$0.03} & 0.54\footnotesize{$\pm$ 0.09} & 0.57\footnotesize{$\pm$ 0.16} & 0.47\footnotesize{$\pm$ 0.09} & 0.51\footnotesize{$\pm$ 0.07} & 0.50\footnotesize{$\pm$ 0.12} & 0.501\footnotesize{$\pm$ 0.016} \\
        \midrule
        Known motif b. & -- & 1.00 \footnotesize{$\pm$ 0.00} & 0.70\footnotesize{$\pm$ 0.03} & 0.99\footnotesize{$\pm$ 0.00} & 0.62\footnotesize{$\pm$0.04} & 1.00\footnotesize{$\pm$ 0.00} & 1.00\footnotesize{$\pm$ 0.00} & 1.00\footnotesize{$\pm$ 0.00} & 1.00\footnotesize{$\pm$ 0.00} & 1.00\footnotesize{$\pm$ 0.00} & 0.890\footnotesize{$\pm$ 0.168} \\
        Known motif c. & -- & 0.92 \footnotesize{$\pm$ 0.00} & 0.56\footnotesize{$\pm$ 0.03} & 0.65\footnotesize{$\pm$ 0.03} & 0.52\footnotesize{$\pm$0.03} & 1.00\footnotesize{$\pm$ 0.00} & 1.00\footnotesize{$\pm$ 0.00} & 0.99\footnotesize{$\pm$ 0.01} & 0.72\footnotesize{$\pm$ 0.09} & 0.63\footnotesize{$\pm$ 0.09} & 0.738\footnotesize{$\pm$ 0.202} \\     
    \bottomrule
    \end{tabular}
        }%
        \vspace{0.1cm}
        \caption[]{Results in terms of AUC of the competing methods on all datasets. The reported errors are standard deviations across $5$ cross-validation (CV) folds (except for the column ``Simulated''). 
        \textbf{Real-world CMV:} Average performance over 
        $5$ CV folds on the \emph{CMV dataset} \citep{emerson2017immunosequencing}. 
        \textbf{Real-world data with implanted signals:} Average performance over
        $5$ CV folds for each 
        of the four datasets. A signal 
        was implanted with a frequency (=witness rate) of 
        $1\%$ or $0.1\%$. Either a single motif (``OM'') 
        or multiple motifs (``MM'') were implanted. 
        \textbf{LSTM-generated data:} Average performance over
        $5$ CV folds for each 
        of the $5$ datasets. In each dataset, a signal 
        was implanted with a frequency of $10\%$,
        $1\%$, $0.5\%$, $0.1\%$, or $0.05\%$, respectively.
        \textbf{Simulated:} Here we report the mean over 18 simulated datasets with implanted signals and varying difficulties (see 
        Tab.~\ref{tab:results_naive} for details). The error reported is the standard deviation of the AUC values across the 18 datasets.
        }%
        \label{tab:results_full}%
    \end{center}%
\end{table}
\endgroup

\textbf{Results.}
In each of the four categories, 
``real-world data'', 
``real-world data with implanted signals'',
``LSTM-generated data'',
and ``simulated immunosequencing data'', 
DeepRC outperforms all competing methods 
with respect to average AUC.
Across categories, the runner-up methods are either the SVM for 
MIL problems with MinMax kernel or the burden test (see Table~\ref{tab:results_full} and Sect.~\ref{sec:detailed_results}). 

\emph{Results on simulated immunosequencing data.}
In this setting the complexity of the implanted signal is in focus and varies 
throughout 18 simulated datasets (see Sect.~\ref{sec:datasets}).
Some datasets are challenging for the methods because
the implanted motif is hidden by noise and others because
only a small fraction of sequences carries the motif, 
and hence have a low witness rate.
These difficulties
become evident by the method called ``known motif binary'', which assumes
the implanted motif is known.
The performance of this method ranges from a
perfect AUC of $1.000$ in several datasets to an 
AUC of $0.532$ in dataset '17' (see Sect.~\ref{sec:detailed_results}).
DeepRC outperforms all other methods with an average AUC of $0.846\pm0.223$, 
followed by the SVM with MinMax kernel with an average AUC of $0.827\pm0.210$ (see Sect.~\ref{sec:detailed_results}).
The predictive
performance of all methods suffers if the signal occurs only in an extremely
small fraction of sequences. %
In datasets, in which only $0.01\%$ of the sequences
carry the motif, all AUC values are below~$0.550$.
\emph{Results on LSTM-generated data.}
On the LSTM-generated data, in which we implanted noisy motifs 
with frequencies of $10\%$, $1\%$, $0.5\%$, $0.1\%$, and $0.05\%$, DeepRC yields almost
perfect predictive performance with an average AUC of $1.000\pm0.001$  
(see Sect.~\ref{sec:detailed_results}~and~\ref{appsec:lstm}). The second best method, 
SVM with MinMax kernel, has a similar predictive performance 
to DeepRC on all datasets but the other 
competing methods have a lower predictive performance 
on datasets with low frequency of the signal ($0.05\%$).
\emph{Results on real-world data with implanted motifs.}
In this dataset category, we used real immunosequences and
implanted single or multiple noisy motifs. Again, 
DeepRC outperforms all other methods with an average AUC of $0.980\pm0.029$,
with the second best method being the burden test with 
an average AUC of $0.883\pm0.170$.
Notably, all methods except for DeepRC have difficulties with noisy
motifs at a frequency of $0.1\%$ 
(see Tab.~\ref{tab:results_milena}). 
\emph{Results on real-world data.}
On the real-world dataset, in which the immune status of persons affected
by the cytomegalovirus has to be predicted, the competing methods yield 
predictive AUCs between $0.515$ and $0.825$ (see Table~\ref{tab:results_full}). We note that this dataset is 
not the exact dataset that was used in  
\citet{emerson2017immunosequencing}.
It differs in pre-processing and also comprises a 
different set of samples and a 
smaller training set due to the nested 5-fold cross-validation procedure,
which leads to a more challenging dataset. 
The best performing 
method is DeepRC with an AUC of $0.831\pm0.002$, followed by
the SVM with MinMax kernel (AUC $0.825\pm0.022$) and the burden
test with an AUC of $0.699\pm0.041$. The top-ranked sequences 
by DeepRC significantly correspond to those detected by 
\citet{emerson2017immunosequencing}, which we tested 
by a Mann-Whitney U-test with the null hypothesis that the
attention values of the sequences detected by \citet{emerson2017immunosequencing} would be equal to the attention 
values of the remaining sequences ($p$-value of $1.3\cdot10^{-93}$). 
The sequence attention values are displayed in Tab.~\ref{tab:emerson_recovered}.

\textbf{Conclusion.}
We have demonstrated how modern Hopfield networks and attention mechanisms
enable successful classification of the immune status of
immune repertoires.
For this task, 
methods have to identify the discriminating
sequences amongst a large set of sequences in an immune repertoire.
Specifically, 
even motifs within those sequences have to be identified.
We have shown that DeepRC, 
a modern Hopfield network and 
an attention mechanism with a fixed query, 
can solve this difficult task
despite the massive number of instances.
DeepRC furthermore outperforms the compared methods
across a range of different experimental conditions.  

\vspace{\cheatspace}
\section*{Broader Impact}
\vspace{\cheatspace}
\textbf{Impact on machine learning and related scientific fields.}
We envision that with (a) the increasing availability of large immunosequencing 
datasets \citep{kovaltsuk2018observed,corrie2018ireceptor, christley2018vdjserver, zhang2020pird, rosenfeld2018immunedb, shugay2018vdjdb}, 
(b) further fine-tuning of ground-truth benchmarking immune receptor 
datasets \citep{weber2019immunesim, olson2019sumrep, marcou2018high},
(c) accounting for 
repertoire-impacting factors such as age, sex, ethnicity, and environment (potential confounding factors), and
(d) increased GPU memory and increased computing power, it 
will be possible to identify discriminating immune receptor motifs for many 
diseases, potentially even for the current SARS-CoV-2 (COVID-19) 
pandemic \citep{Raybould_cov-abdb_2020,Minervina_tcr_2020, Galson_bcr_2020}. 
Such results would greatly benefit ongoing research on antibody and 
TCR-driven immunotherapies and immunodiagnostics as well as rational 
vaccine design \citep{brown2019augmenting}.

In the course of this development,
the experimental verification and interpretation of machine-learning-identified motifs could receive additional focus,
as for most of the sequences within a repertoire the corresponding antigen is 
unknown.
Nevertheless, recent technological breakthroughs in high-throughput 
antigen-labeled immunosequencing are beginning to generate large-scale antigen-labeled 
single-immune-receptor-sequence data thus resolving this longstanding problem \citep{setliff2019high}.

From a machine learning perspective, the successful application of DeepRC on immune repertoires with their large number of instances per bag might encourage the application of modern Hopfield networks and attention mechanisms on new, previously unsolved or unconsidered, datasets and problems.

\textbf{Impact on society.}
If the approach proves itself successful, it could lead to faster testing 
of individuals for their immune status w.r.t. a range of diseases based on 
blood samples.
This might motivate changes in the pipeline of diagnostics 
and tracking of diseases, e.g. automated testing of the immune status in 
regular intervals.
It would furthermore make the collection and screening 
of blood samples for larger databases more attractive.
In consequence, the improved testing of immune statuses might identify 
individuals that do not have a working immune response towards certain 
diseases to government or insurance companies, which could then push 
for targeted immunisation of the individual.
Similarly to compulsory 
vaccination, such testing for the immune status could be made compulsory 
by governments, possibly violating privacy or personal self-determination
in exchange for increased over-all health of a population.

Ultimately, if the approach proves itself successful, the insights gained from the screening of individuals that have successfully developed resistances against specific diseases could lead to faster targeted immunisation, once a certain number of individuals with resistances can be found.
This might strongly decrease the harm done by e.g. pandemics and lead to a change in the societal perception of such diseases.

\textbf{Consequences of failures of the method.}
As common with methods in machine learning, 
potential danger lies in the possibility that users rely too much on 
our new approach and use it without reflecting on the outcomes.
However, the full pipeline in which our method would be used includes wet lab tests after its application,
to verify and investigate the results, gain insights, and possibly derive treatments.
Failures of the proposed method would lead to unsuccessful wet lab validation and
negative wet lab tests.
Since the proposed algorithm does not directly suggest treatment or therapy,
human beings are not directly at risk of being treated with a harmful therapy.
Substantial wet lab and in-vitro testing and would indicate wrong decisions by the system.

\textbf{Leveraging of biases in the data and potential discrimination.}
As for almost all machine learning methods, confounding factors, such as age or sex, could be used for classification.
This, might lead to biases in predictions or uneven predictive performance
across subgroups. As a result, failures in the wet lab would occur (see paragraph above). Moreover, insights into the relevance of the confounding factors could be gained, leading to possible therapies or counter-measures concerning said factors.

Furthermore, the amount of data available with respec to
relevant confounding factors could lead to better or worse performance 
of our method.
E.g. a dataset consisting mostly of data from individuals within a specific age 
group might yield better performance for that age group,
possibly resulting in better or exclusive treatment methods for that specific 
group.
Here again, the application of DeepRC would be followed by in-vitro testing and 
development of a treatment,
where all target groups for the treatment have to be considered accordingly.

\section*{Availability}
All datasets and code is available at \url{https://github.com/ml-jku/DeepRC}.
The CMV dataset is publicly available at \url{https://clients.adaptivebiotech.com/pub/Emerson-2017-NatGen}.

\section*{Acknowledgments}
The ELLIS Unit Linz, the LIT AI Lab and the 
Institute for Machine Learning are supported by
the Land Oberösterreich,
LIT grants DeepToxGen (LIT-2017-3-YOU-003),
and AI-SNN (LIT-2018-6-YOU-214),	
the Medical Cognitive Computing Center (MC3),
Janssen Pharmaceutica,
UCB Biopharma,
Merck Group,
Audi.JKU Deep Learning Center, Audi Electronic Venture GmbH,
TGW,
Primal,
Silicon Austria Labs (SAL),
FILL,
EnliteAI,
Google Brain,
ZF Friedrichshafen AG,
Robert Bosch GmbH,
TÜV Austria,
DCS,
and the NVIDIA Corporation. 
Victor Greiff (VG) and Geir Kjetil Sandve (GKS) are supported by The Helmsley 
Charitable Trust (\#2019PG-T1D011, to VG), UiO World-Leading Research 
Community (to VG), UiO:LifeSciences Convergence Environment Immunolingo 
(to VG and GKS), EU Horizon 2020 iReceptorplus (\#825821, to VG) and 
Stiftelsen Kristian Gerhard Jebsen (K.G. Jebsen Coeliac Disease Research 
Centre, to GKS). 

\newpage

\chapter*{Appendix}

In the following, the appendix to the
paper ``Modern Hopfield Networks and Attention for Immune Repertoire Classification'' is presented.
Here we provide details on DeepRC,
the compared methods, and the experimental setup and results.
Furthermore,
the generation of the immune repertoire classification data using an LSTM network,
the interpretation of DeepRC and the extraction of found motifs,
and the ablation study using different variants of DeepRC
are described.

\newpage

\renewcommand{\contentsname}{\Large Contents of Appendix \vspace{-1.5cm}}
\setcounter{tocdepth}{2}
\begingroup
\let\clearpage\relax
\vspace{-2cm} 
\tableofcontents

\vspace{-1.8cm} 

\renewcommand{\listfigurename}{\Large List of figures 
\vspace{-1.8cm}}
\listoffigures

\vspace{-1.8cm} 

\renewcommand{\listtablename}{\Large List of tables \vspace{-1.8cm}}
\listoftables

\endgroup

\renewcommand{\thesection}{A\arabic{section}}
\renewcommand{\thefigure}{A\arabic{figure}}
\renewcommand{\thetable}{A\arabic{table}}
\renewcommand{\theequation}{A\arabic{equation}}

\setcounter{theorem}{0}
\setcounter{section}{0}
\setcounter{table}{0}
\setcounter{figure}{0}
\setcounter{equation}{0}

\newpage
\chapter{Immune Repertoire Classification}
\chapterauthor{Michael Widrich \quad
    Bernhard Sch\"{a}fl \quad 
    Milena Pavlovi{\'c} \quad
    Geir Kjetil Sandve \quad 
    Sepp Hochreiter \quad 
    Victor Greiff \quad
    G\"{u}nter Klambauer}
    
\section{Introduction}
In Section~\ref{sec:deeprc_details} we provide details 
on the architecture of DeepRC, in Section~\ref{sec:datasets} 
we present details on the datasets, in Section~\ref{subsec:methods_compared} we explain the 
methods that we compared, in Section~\ref{sec:hyperparams} we
elaborate on the hyperparameters and their selection process. 
Then, in Section~\ref{sec:detailed_results} we present 
detailed results for each dataset category in tabular form, 
in Section~\ref{appsec:lstm} we provide information on 
the LSTM model that was used to generate antibody sequences,
in Section~\ref{appsec:integrated_gradients} we show 
how DeepRC can be interpreted, in Section~\ref{sec:attention_cmv}
we show the correspondence of previously identified 
TCR sequences for CMV immune status with attention values 
by DeepRC, and finally we present variations and an ablation 
study of DeepRC in Section~\ref{sec:deeprc_variations}.

\section{DeepRC implementation details}
\label{sec:deeprc_details}
\label{sec:deeprc_implementation_details}
\begin{figure}[htp]
    \begin{center}
    \includegraphics[width=0.8\textwidth]{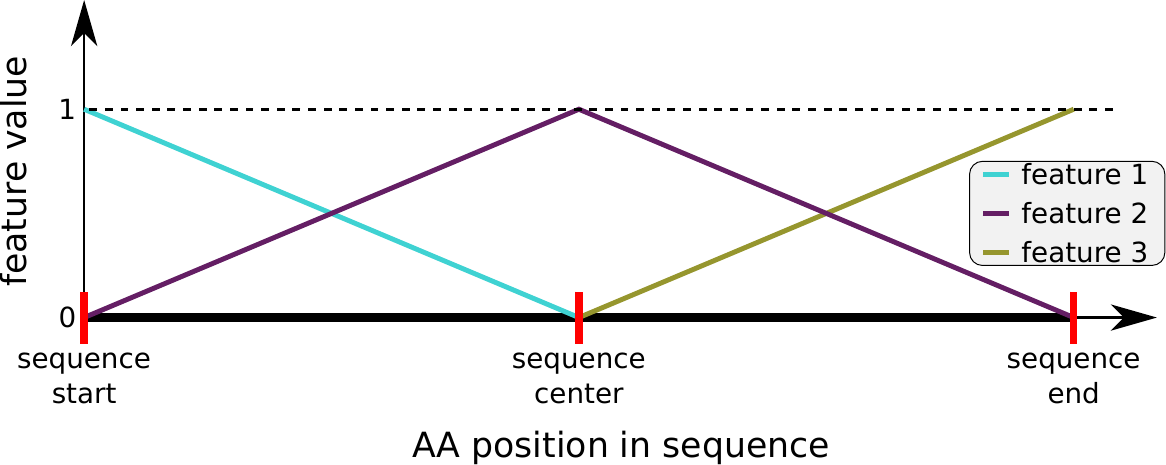}\\
   \caption[Position encoding]{We use 3 input features with values in range $[0,1]$ to encode the relative position of each AA in a sequence with respect to the sequence.
   ``feature 1'' encodes if an AA is close to the sequence start,
   ``feature 2'' to the sequence center, and 
   ``feature 3'' to the sequence end.
   For every position in the sequence, the values of all three features sum up to $1$.
   \label{fig:positional_encoding}}
   \end{center}
\end{figure}

\paragraph{Input layer.} For the input layer of the CNN, 
the characters in the input sequence, i.e. the amino acids (AAs),
are encoded in a one-hot vector of length $20$.
To also provide information about the position of an AA in the sequence, we add 
$3$ additional input features with values in range $[0, 1]$ to encode the position 
of an AA relative to the sequence.
These $3$ positional features encode whether the AA is located at the beginning, 
the center, or the end of the sequence, respectively, as shown in 
Figure~\ref{fig:positional_encoding}.
We concatenate these $3$ positional features with the one-hot vector of AAs,
which results in a feature vector of size $23$ 
per sequence position.
Each repertoire, now represented as a bag of feature vectors, 
is then normalized to %
unit variance.
Since the cytomegalovirus dataset (\emph{CMV dataset}) provides sequences with an associated 
abundance value per sequence,
which is the number of occurrences of a sequence in a repertoire,
we incorporate this information into the input of DeepRC.
To this end, the one-hot AA features of a sequence are 
multiplied by a scaling factor of $\log(c_a)$
before normalization,
where $c_a$ is the abundance of a sequence.
We feed the sequences with $23$ features per position into the CNN.
Sequences of different lengths were zero-padded to the maximum sequence length
per batch at the sequence ends.

\paragraph{1D CNN for motif recognition.}
In the following,
we describe how DeepRC identifies patterns in the individual sequences and 
reduces each sequence in the input object to a fixed-size feature vector.
DeepRC employs 1D convolution layers to extract
patterns, where trainable weight kernels are convolved over the sequence positions.
In principle, also recurrent neural networks (RNNs) or transformer networks could be used
instead of 1D CNNs, however, (a) the computational complexity of the network 
must be low to be able to process millions of sequences for a single update. Additionally, 
(b) the learned network should be able to provide insights in the recognized patterns in form
of motifs. Both properties (a) and (b) are fulfilled by 1D convolution operations that are used
by DeepRC. 

We use one 1D CNN layer \citep{hu2014convolutional}
with SELU activation functions \citep{klambauer2017self}
to identify
the relevant patterns in the input sequences with a computationally light-weight operation.
The larger the kernel size,
the more surrounding sequence positions are taken into account,
which influences the length of the motifs that can be extracted.
We therefore adjust the kernel size 
during hyperparameter search.
In prior works \citep{ostmeyer2019biophysicochemical},  
a k-mer size of $4$ yielded good predictive performance,
which could indicate that a kernel size in the range of $4$ may 
be a proficient choice.
For $d_v$ trainable kernels, this produces a feature vector 
of length $d_v$ at each sequence position. %
Subsequently, global max-pooling over all sequence positions of a sequence reduces the sequence-representations $\Bz_i$ to vectors of the fixed length $d_v$. 
Given the challenging size of the input data per repertoire,
the computation of the CNN activations and weight updates is 
performed using 16-bit floating point values.
A list of hyperparameters evaluated for DeepRC is given in 
Table~\ref{tab:deeprc_settings}.
A comparison of RNN-based and CNN-based sequence embedding for motif recognition in a smaller experimental setting is given in Sec.~\ref{sec:deeprc_variations}.

\paragraph{Regularization.}
We apply random and attention-based subsampling of repertoire sequences to
reduce over-fitting and decrease computational effort.
During training, each repertoire is subsampled to $10,000$ input sequences,
which are randomly drawn from the respective repertoire.
This can also be interpreted as random drop-out \citep{hinton2012improving} on 
the input sequences or attention weights.
During training and evaluation,
the attention weights computed by the attention network are furthermore used to 
rank the input sequences.
Based on this ranking, the repertoire is reduced to the $10\%$ of sequences 
with the highest attention weights.
These top $10\%$ of sequences are then used to compute the weight updates and
the prediction for the repertoire.
Additionally, one might employ further regularization techniques,
which we only partly investigated further in a smaller experimental setting in Sec.~\ref{sec:deeprc_variations} due to high computational demands.
Such regularization techniques include
$l1$ and $l2$ weight decay,
noise in the form of random AA permutations in the input sequences,
noise on the attention weights,
or random shuffling of sequences between repertoires that belong to the negative class.
The last regularization technique assumes that the sequences in positive-class repertoires 
carry a signal, such as an AA motif corresponding to an immune response,
whereas the sequences in negative-class repertoires do not.
Hence, the sequences can be shuffled randomly between negative class repertoires without obscuring the 
signal in the positive class repertoires.

\paragraph{Hyperparameters.} 
For the hyperparameter search of DeepRC for the category ``simulated 
immunosequencing data'',
we only conducted a full hyperparameter 
search on the more difficult datasets with motif implantation probabilities below $1\%$,
as described in Table~\ref{tab:deeprc_settings}.
This process was repeated for all $5$ folds of the 5-fold cross-validation (CV) and 
the average score on the $5$ test sets constitutes the final score of a method.

Table~\ref{tab:deeprc_settings} provides an overview of the hyperparameter 
search,
which was conducted as a grid search
for each of the datasets in a nested 5-fold CV procedure,
as described in Section~\ref{subsec:methods_compared}.

\paragraph{Computation time and optimization.}
We took measures on the implementation level to address the high computational demands, especially GPU memory consumption, in order to make the large number of experiments feasible.
We train the DeepRC model with a small batch size of $4$ samples
and perform computation of inference and updates of the 1D CNN using 16-bit floating point values.
The rest of the network is trained using 32-bit floating point values.
The Adam parameter for numerical stability was therefore
increased from the default value of $\epsilon=10^{-8}$ to $\epsilon=10^{-4}$.
Training was performed on various GPU types,
mainly \texttt{NVIDIA RTX 2080 Ti}.
Computation times were highly dependent on the number of sequences in the repertoires
and the number and sizes of CNN kernels.
A single update on an \texttt{NVIDIA RTX 2080 Ti} GPU took approximately $0.0129$ to $0.0135$ seconds,
while requiring approximately $8$ to $11$ GB GPU memory.
Taking these optimizations and 
GPUs with larger memory ($\geq16$ GB) into account,
it is already possible to train DeepRC,
possibly with multi-head attention and a larger network architecture,
on larger datasets (see Sec.~\ref{sec:deeprc_variations}).
Our network implementation is based on \texttt{PyTorch 1.3.1} \citep{paszke2019pytorch}.

\paragraph{Incorporation of additional inputs and metadata.}
Additional metadata in the form of sequence-level or repertoire-level features
could be incorporated into the input
via concatenation with the feature vectors that result from
taking the maximum of the 1D CNN outputs w.r.t. the sequence positions.
This has the benefit that
the attention mechanism and output network can utilize the sequence-level or repertoire-level features for their predictions.
Sparse metadata or metadata that is only available during training
could be used as auxiliary targets
to incorporate the information via gradients into the DeepRC model.

\paragraph{Limitations.}
The current methods are mostly limited by 
computational complexity, since
both hyperparameter and model selection is 
computationally demanding.
For hyperparameter selection, 
a large number of hyperparameter settings have to be evaluated. 
For model selection, a single repertoire requires 
the propagation of many thousands of sequences
through a neural network and keeping those quantities 
in GPU memory in order to perform the attention mechanism and weight update.
Thus, increased GPU memory would significantly boost 
our approach. Increased computational power would 
also allow for more advanced architectures and 
attention mechanisms, which may further improve 
predictive performance.
Another limiting factor is over-fitting of the model due to the currently relatively small number of samples (bags) in real-world immunosequencing datasets 
in comparison to the large number of instances per bag and 
features per instance.

\section{Datasets}\label{sec:datasets}
We aimed at constructing immune repertoire classification 
scenarios with varying degree of realism and 
difficulties in order to compare and analyze
the suggested machine learning methods.
To this end, we either use
simulated or experimentally-observed immune receptor sequences and we implant signals,
which are sequence motifs \citep{akbar2019compact, weber2019immunesim}, into sequences of repertoires of the positive class. It has been shown previously that interaction of immune receptors with antigens occur via short sequence stretches \citep{akbar2019compact}. Thus, implantation of short motif sequences simulating an immune signal is biologically meaningful.
Our benchmarking study comprises four different categories of datasets:
(a) Simulated immunosequencing data with implanted signals (where the signal is defined as sets of motifs), 
(b) LSTM-generated immunosequencing data with implanted signals,
(c) real-world immunosequencing data with implanted signals, and 
(d) real-world immunosequencing data.
Each of the first three categories consists of multiple datasets with varying 
difficulty depending on the type of the implanted signal and the ratio of sequences with the implanted signal.
The ratio of sequences with the implanted signal,
where each sequence carries at most 1 implanted signal,
corresponds to the \emph{witness rate} (WR).
We consider binary classification tasks to simulate the immune status
of healthy and diseased individuals. We randomly generate immune repertoires
with varying numbers of sequences, where we implant sequence motifs in the 
repertoires of the diseased individuals, i.e. the positive class. The sequences of a repertoire are 
also randomly generated by different procedures (detailed below).
Each sequence is composed of $20$ different characters, corresponding to amino acids, 
and has an average length of $14.5$ AAs.

\subsection{Simulated immunosequencing data}
In the first category,
we aim at investigating the impact of the signal frequency, i.e. the WR, and the signal 
complexity on the performance of the different methods.
To this end, we created $18$ datasets,
whereas each dataset contains a large number of repertoires
with a large number of random AA sequences per repertoire.
We then implanted signals in the AA sequences of the positive class repertoires,
where the $18$ datasets differ in frequency and complexity of the implanted signals.
In detail, the AAs were sampled randomly independent of their respective position in the sequence,
while the frequencies of AAs, 
distribution of sequence lengths, 
and distribution of the number of sequences per repertoire,
i.e. the number of instances per bag, are 
following the respective distributions observed in the real-world \emph{CMV dataset}~\citep{emerson2017immunosequencing}.
For this, we first sampled the number of sequences for a repertoire from 
a Gaussian $\mathcal N(\mu=316k, \sigma=132k)$ distribution and 
rounded to the nearest positive integer.
We re-sampled if the size was below $5k$. 
We then generated random sequences of AAs with a length of $\mathcal N(\mu=14.5, \sigma=1.8)$,
again rounded to the nearest positive integers.
Each simulated repertoire was then randomly assigned to
either the positive or negative class,
with $2,500$ repertoires per class.
In the repertoires assigned to the positive class,
we implanted motifs with an average length of $4$ AAs,
following the results of the experimental analysis of antigen-binding motifs in antibodies and T-cell receptor sequences by \citep{akbar2019compact}.
We varied the characteristics of the implanted motifs for each of the $18$ datasets with respect to the following parameters:
(a) $\rho$, the probability of a motif being implanted in a sequence of a positive repertoire, 
i.e. the average ratio of sequences containing the motif, which is the witness rate.  
(b) The number of wildcard positions in the motif.
A wildcard position contains a random AA, which is randomly sampled for each sequence.
Wildcard positions are located in the center of the implanted motif.
(c) The number of deletion positions in the implanted motif.
A deletion position has a probability of $0.5$ of being removed from the motif. 
Deletion positions are located in the center of the implanted motifs. 

In this way, we generated $18$ different datasets of variable 
difficulty containing in total roughly $28.7$ billion sequences. 
See Table~\ref{tab:lstm_simulations} for an overview of the properties 
of the implanted motifs in the $18$ datasets.

\subsection{LSTM-generated data}\label{subsec:lstm_generated_data}
In the second dataset category,
we investigate the impact of the signal frequency and complexity
in combination with more plausible immune receptor sequences
by taking into account the positional AA distributions
and other sequence properties.
To this end, we trained an LSTM \citep{hochreiter1997long} in a standard 
next character prediction \citep{graves2013generating}
setting to create AA sequences with properties similar to experimentally observed immune receptor sequences.

In the first step, the LSTM model  
was trained on all immuno-sequences in the \emph{CMV dataset} 
\citep{emerson2017immunosequencing} that contain valid information 
about sequence abundance and have a known CMV label.
Such an LSTM model is able to capture various properties of the sequences,
including position-dependent probability distributions and
combinations, relationships, and order of AAs.
We then used the trained LSTM model to generate $1,000$ repertoires
in an autoregressive fashion,
starting with a start sequence that was randomly 
sampled from the trained-on dataset.
Based on a visual inspection of the frequencies of 4-mers (see Section~\ref{appsec:lstm}),
the similarity of LSTM generated sequences and real sequences
was deemed sufficient for the purpose of generating the AA sequences
for the datasets in this category.
Further details on LSTM training and repertoire generation are given in Section~\ref{appsec:lstm}.

After generation,
each repertoire was assigned to either the positive or negative class,
with $500$ repertoires per class.
We implanted motifs of length $4$ with varying properties
in the center of the sequences of the positive class
to obtain $5$ different datasets.
Each sequence in the positive repertoires has a probability 
$\rho$ to carry the motif,
which was varied throughout $5$ datasets
and corresponds to the WR (see Table~\ref{tab:lstm_simulations}).
Each position in the motif has a probability of $0.9$ to be implanted
and consequently a probability of $0.1$ that the original AA in the sequence remains,
which can be seen as noise on the motif.

\begin{table}[htp]%
    \begin{center}%
        \begin{tabular}{laca}%
            \toprule
                                   & Simulated                   & LSTM gen. & Real-world\\
            \midrule
            seq. per bag & $N(316k, 132k)$ & $N (285k, 156k)$ & $10k$\\
            repertoires              & $5,000$                             & $1,000$ & $1,500$\\
            motif noise          & $0\%$                                 & $10\%$ & $*$  \\
            wildcards            & $\{0;1;2\}$                         & $0$ & $0$ \\
            deletions            & $\{0;1\}$                           & $0$ & $0$ \\
            mot. freq. $\rho$     & $\{1; 0.1;$                      & $\{10;1;0.5;$ & $\{1;0.1\}$ \\
            (in $\%$)            & $0.01\}$                           & $0.1;0.05\}$ &  \\
            \bottomrule
        \end{tabular}%
        \caption[Properties of simulated repertoires,
        variations of motifs, 
        and motif frequencies]{Properties of simulated repertoires,
        variations of motifs, 
        and motif frequencies, i.e. the witness rate, for 
        the datasets in categories 
        ``simulated immunosequencing data'',
        ``LSTM-generated data'',
        and ``real-world data with implanted signals''.
        Noise types for $*$ are explained in paragraph ``real-world data with implanted signals''.}%
        \label{tab:lstm_simulations}%
    \end{center}%
\end{table}

\subsection{Real-world data with implanted signals}
In the third category, we implanted signals into experimentally obtained 
immuno-sequences, where we considered $4$ dataset variations.
Each dataset consists of $750$ repertoires for each of the two classes, 
where each repertoire consists of $10k$ sequences. In this way, we 
aim to simulate datasets with a \emph{low sequencing coverage}, which
means that only relatively few sequences per repertoire are available.
The sequences were randomly sampled from healthy (CMV negative) individuals from the \emph{CMV dataset} (see below paragraph for explanation).
Two signal types were considered:
(a) \textbf{One signal with one motif.}
The AA motif \texttt{LDR} was implanted in a certain fraction 
of sequences. The pattern is randomly altered at one of the 
three positions with probabilities $0.2$, $0.6$, and $0.2$, 
respectively. 
(b) \textbf{One signal with multiple motifs.}
One of the three possible motifs \texttt{LDR}, \texttt{CAS}, and
\texttt{GL-N} was implanted with equal probability. Again, the motifs 
were randomly altered before implantation. 
The AA motif \texttt{LDR} changed as described above. 
The AA motif \texttt{CAS} was altered at the second position with 
probability $0.6$ and with probability $0.3$ at the first position. 
The pattern \texttt{GL-N}, where \texttt{-} denotes a gap location,
is randomly altered at the first position with probability $0.6$ and 
the gap has a length of $0$, $1$, or $2$ AAs with equal probability.

Additionally, the datasets differ in the values for $\rho$, the 
average ratio of sequences carrying a signal, 
which were chosen as $1\%$ or $0.1\%$.
The motifs were implanted at positions $107$, $109$, and $114$ according to the IMGT numbering scheme for immune receptor sequences \citep{lefranc2003imgt} 
with probabilities $0.3$, $0.35$ and $0.2$, respectively.
With the remaining $0.15$ chance, 
the motif is implanted at any other sequence position.
This means that the motif occurrence in the simulated sequences is biased towards the middle of the sequence.

\subsection{Real-world data: CMV dataset}
We used a real-world dataset of $785$ repertoires,
each of which containing between $4,371$ to $973,081$ (avg. $299,319$) TCR sequences
with a length of $1$ to $27$ (avg. $14.5$) AAs, 
originally collected and provided by \citet{emerson2017immunosequencing}.
$340$ out of $785$ repertoires 
were labelled as positive for cytomegalovirus (CMV) serostatus, which we consider
as the positive class, $420$ repertoires with negative CMV serostatus, considered
as negative class, and $25$ repertoires with unknown status.
We changed the number of 
sequence counts per repertoire from $-1$ to $1$ 
for $3$ sequences.
Furthermore, 
we exclude a total of $99$ repertoires with unknown CMV status or unknown 
information about the sequence abundance within a repertoire, 
reducing the dataset for our analysis to $686$ repertoires, 
$312$ of which with positive and $374$ with negative CMV status.

\subsection{Comparison to other MIL datasets}
We give a non-exhaustive overview of previously considered
MIL datasets and problems in Table~\ref{tab:mil_datasets}. To our knowledge
the datasets considered in this work pose the most
challenging MIL problems with respect to the number of instances
per bag (column 5). 

\begin{table}[htbp]
    \begin{center}%
    \rotatebox{90}{\begin{varwidth}{\textheight}
    \begin{center}%
    \resizebox{0.870\textheight}{!}{%
        \begin{tabular}{lacarac}%
            \toprule
            Dataset & Total number & Total number & Approx. number of & Avg. number of & Source & Dataset\\
             & of bags & of instances & features per instance & instances per bag &  & reference\\
            \midrule
            Simulated immuno- & 5,000 & 1,597,024,310 & 14.5x20 AA sequence & 316,000 & this work & \\
            sequencing data (ours) & & x 18 datasets & & & & \\
             & &  & & & & \\
            LSTM-generated data (ours) & 1,000 & 304,825,671 & 14.5x20 AA sequence & 285,000 & this work & \\
             & & x 5 datasets & & & & \\
             & &  & & & & \\
            Real-world data with  & 1,500 & 14,715,421 & 14.5x20 AA sequence & 10,000 & this work & \\
            implanted signals (ours) & & x 4 datasets & & & & \\
             & &  & & & & \\
            CMV (pre-processed by us) & 785 & 234,965,729 & 14.5x20 AA sequence & 299,000 & this work & \citet{emerson2017immunosequencing}\\
             & &  & & & & \\
            MNIST bags & 50--500 & 500--50,000 & 28x28x1 image & 100 & \citet{ilse2018attention} & \\
             & &  & & & & \\
            Breast Cancer & 58 & approx. 39,000 & 32x32x3 H\&E image & 672 & \citet{ilse2018attention} & \citet{gelasca2008evaluation} \\
             & &  & & & & \\
            Basal cell carcinomas & 820 & 7,588,767 & 1024x1024x3 H\&E image & 9,056 & \citet{kimeswenger2019detecting} & \\
             & &  & & & & \\
            Birds & 548 & 10,232 & 38 & 9 & \citet{ruiz2018knnmiml} & \citet{briggs2012rank}\\
             & &  & & & & \\
            Scene & 2,000 & 18,000 & 15 & 9 & \citet{ruiz2018knnmiml} & \citet{zhang2007multi}\\
             & &  & & & & \\
            Reuters & 2,000 & 7,119 & 243 & 4 & \citet{ruiz2018knnmiml} & \citet{sebastiani2002machine}\\
             & &  & & & & \\
            CK+ & 430 & 7,915 & 4,391 & 18 & \citet{ruiz2018knnmiml} & \citet{lucey2010extended}\\
             & &  & & & & \\
            UniProt (Geobacter sulfurreducens) & 379 & 1,250 & 216 & 3 & \citet{ruiz2018knnmiml} & \citet{wu2014genome}\\
             & &  & & & & \\
            MODIS (aerosol data) & 1,364 & 136,400 & 12 & 100 & \citet{uriot2019learning} & \url{https://aeronet.gsfc.nasa.gov}\\
             & &  & & & & \\
            MISR1 (aerosol data) & 800 & 80,000 & 16 & 100 & \citet{uriot2019learning} & \url{https://aeronet.gsfc.nasa.gov}\\
             & &  & & & & \\
            MISR2 (aerosol data) & 800 & 80,000 & 12 & 54 & \citet{uriot2019learning} & \url{https://aeronet.gsfc.nasa.gov}\\
             & &  & & & & \\
            CORN (crop yield) & 525 & 52,500 & 92 & 100 & \citet{uriot2019learning} & \url{https://aeronet.gsfc.nasa.gov}\\
             & &  & & & & \\
            WHEAT (crop yield) & 525 & 52,500 & 92 & 100 & \citet{uriot2019learning} & \url{https://aeronet.gsfc.nasa.gov}\\
             & &  & & & & \\
            \bottomrule
        \end{tabular}%
        }
        \parbox{0.87\textheight}{\caption[Overview of MIL datasets and their characteristics]{MIL datasets with their numbers of bags and numbers of instances. ``total number of instances'' refers to the total number of instances in the dataset.
        The simulated and real-world immunosequencing datasets considered in this work contain a by orders of magnitudes larger number of instances per bag than MIL datasets that were considered by machine learning methods up to now.}%
        \label{tab:mil_datasets}}%
    \end{center}%
    \end{varwidth}%
    }%
    \end{center}%
\end{table}

\clearpage
\section{Compared methods}\label{subsec:methods_compared}
We evaluate and compare the performance of 
DeepRC against a set of machine learning methods that
serve as baseline, were suggested, or 
can readily be adapted to immune repertoire 
classification. In this section, we describe these compared 
methods.

\subsection{Known motif}
This method serves as an estimate for the achievable classification performance 
using prior knowledge about which motif was implanted. 
Note that this does not necessarily lead to perfect predictive performance
since motifs are implanted with a certain amount of noise
and could also be present in the negative class by chance.
The \emph{known motif} method counts how often the known implanted motif occurs 
per sequence for each repertoire and uses this count to rank the repertoires.
From this ranking, 
the Area Under the receiver operator Curve (AUC) is computed as performance measure.
Probabilistic AA changes in the known motif are not considered for this count,
with the exception of gap positions.
We consider two versions of this method:
(a) \textbf{Known motif binary:} counts the occurrence of the known motif in a sequence
and (b) \textbf{Known motif continuous:} counts the maximum number of overlapping AAs between the 
known motif and all sequence positions,
which corresponds to a convolution operation with a binary kernel followed by max-pooling.
Since the implanted signal is not known in the experimentally obtained \emph{CMV dataset},
this method cannot be applied to this dataset.

\subsection{Support Vector Machine (SVM)}
The Support Vector Machine (SVM) approach uses a fixed mapping from a bag 
of sequences to the corresponding k-mer counts. 
The function $h_{\text{kmer}}$ maps each sequence $s_i$ to a
vector representing the occurrence of k-mers in the sequence. To avoid confusion 
with the sequence-representation obtained from the CNN layers of 
DeepRC, we denote $\Bu_i=h_{\text{kmer}}(s_i)$, which is analogous to $\Bz_i$. 
Specifically, $u_{im}= \left(h_{\text{kmer}}(s_i)\right)_m = \#\{p_m \in s_i\}$, where $\# \{p_m \in s_i\}$
denotes how often the k-mer pattern $p_m$ occurs in sequence $s_i$. Afterwards, 
average-pooling is applied to obtain $\Bu=1/N \sum_{i=1}^N \Bu_i$, 
the \emph{k-mer representation} of the input object $X$. For 
two input objects $X^{(n)}$ and $X^{(l)}$ with representations
$\Bu^{(n)}$ and $\Bu^{(l)}$, respectively, we implement the \emph{MinMax kernel}~\citep{ralaivola2005graph} as follows:
\begin{equation}
    \begin{split}
        k(X^{(n)}, X^{(l)})&= k_{\mathrm{MinMax}} (\Bu^{(n)},\Bu^{(l)}) \\
        &= \frac{\sum_{m=1}^{d_u} \min(u^{(n)}_{m},u^{(l)}_{m})}{\sum_{m=1}^{d_u}  \max(u^{(n)}_{m},u^{(l)}_{m})},
    \end{split}
\end{equation}
where $u^{(n)}_{m}$ is the $m$-th element of the vector $u^{(n)}$. 
The \emph{Jaccard kernel}~\citep{levandowsky1971distance} is identical to the MinMax kernel
except that it operates on binary $\Bu^{(n)}$.
We used a standard C-SVM, as introduced by \citet{cortes1995support}. 
The corresponding hyperparameter $C$ is optimized by random search. The settings of the 
full hyperparameter search as well as the respective value ranges are 
given in \tablename~\ref{tab:svm_settings}.

\subsection{K-Nearest Neighbor (KNN)}
The same \emph{k-mer representation} of a repertoire, as introduced above for the SVM baseline,
is used for the K-Nearest Neighbor (KNN) approach. As this method clusters samples 
according to distances between them, the previous kernel definitions cannot be applied 
directly. It is therefore necessary to transform the MinMax as well as the Jaccard 
kernel from similarities to distances by constructing 
the following \citep{levandowsky1971distance}:
\begin{equation}
    \begin{split}
        d_{\mathrm{MinMax}} (\Bu^{(n)},\Bu^{(l)}) &= 1 - k_{\mathrm{MinMax}} (\Bu^{(n)},\Bu^{(l)}), \\
        d_{\mathrm{Jaccard}} (\Bu^{(n)},\Bu^{(l)}) &= 1 - k_{\mathrm{Jaccard}} (\Bu^{(n)},\Bu^{(l)}).
    \end{split}
\end{equation}
The amount of neighbors is treated as the hyperparameter and optimized by an 
exhaustive grid search. The settings of the full hyperparameter search as well 
as the respective value ranges are given in \tablename~\ref{tab:knn_settings}.

\subsection{Logistic regression}
We implemented logistic regression on the \emph{k-mer 
representation} $\Bu$ of an immune repertoire. The model
is trained by gradient descent using the Adam optimizer~\citep{kingma2014adam}.
The learning rate is treated as the hyperparameter and optimized by grid search.
Furthermore, we explored two regularization settings 
using combinations of $l1$ and $l2$ weight decay.
The settings of the full hyperparameter search as well as the respective value ranges
are given in \tablename~\ref{tab:logistic_regression_settings}.

\subsection{Burden test} 
We implemented a burden test \citep{emerson2017immunosequencing,li2008methods,wu2011rare} in a machine learning
setting. The burden test first identifies sequences or k-mers that are associated
with the individual's class, i.e., immune status, and then calculates a burden 
score per individual. Concretely, for each k-mer or sequence, the phi coefficient
of the contingency table for absence or presence and positive or negative immune status
is calculated. Then, $J$ k-mers or sequences with the highest phi coefficients 
are selected as the set of associated k-mers or sequences. $J$ is a hyperparameter that 
is selected on a validation set. Additionally, we consider the type of input 
features, sequences or k-mers, as a hyperparameter. For inference, a burden score
per individual is calculated as the sum of associated k-mers or sequences it carries. 
This score is used as raw prediction and to rank the individuals.
Hence, we have extended the burden test by \citet{emerson2017immunosequencing} 
to k-mers and to adaptive thresholds that are adjusted on a validation set. 

\subsection{Logistic MIL (Ostmeyer et al)}
The logistic multiple instance learning (MIL) approach for immune 
repertoire classification \citep{ostmeyer2019biophysicochemical} applies a logistic 
regression model to each k-mer representation in a bag.
The resulting scores are then 
summarized by max-pooling to obtain a prediction for the bag.
Each amino acid of each k-mer is represented by $5$ features, the so-called 
Atchley factors~\citep{atchley2005solving}. As k-mers of length $4$ are used, 
this gives a total of $4\times5=20$ features.
One additional feature per 4-mer is added, which represents the relative 
frequency of this 4-mer with respect to its containing bag, resulting in 
$21$ features per 4-mer.
Two options for the relative frequency feature exist, which are
(a) whether the frequency of the 4-mer (``4MER'') or
(b) the frequency of the sequence in which the 4-mer appeared (``TCR\textbeta'') is used.
We optimized the learning rate, batch size, and early stopping parameter 
on the validation set. The settings of the full hyperparameter search as
well as the respective value ranges are given in
\tablename~\ref{tab:mil_settings}.

\clearpage
\section{Hyperparameter selection}
\label{sec:hyperparams}
For all competing methods a hyperparameter search was performed,
for which we split each 
of the $5$ training sets into an inner training set and 
inner validation set.
The models were trained on the inner training set and 
evaluated on the inner validation set.
The model with the highest AUC score on the inner validation 
set is then used to 
calculate the score on the respective test set.
Here
we report the hyperparameter sets and search strategy that is
used for all methods. 

\paragraph{DeepRC.}
The set of hyperparameters of DeepRC is shown in 
Table~\ref{tab:deeprc_settings}. These hyperparameter
combinations are adjusted via a grid search procedure.

\begin{table}[htp]%
    \begin{center}%
        \begin{tabular}{lc}%
            \toprule
            learning rate & $10^{-4}$ \\
            number of kernels ($d_v$) & $\{8;16;32;64^*;128^*;256^*\}$ \\
            number of CNN layers & $\{1\}$ \\
            number of layers in key-NN & $\{2\}$ \\
            number of units in key-NN & $\{32\}$ \\
            kernel size & $\{5;7;9\}$ \\
            subsampled seqences & $10,000$ \\
            batch size & $4$ \\
            \bottomrule
        \end{tabular}%
        \caption[DeepRC hyperparameter search space]{\textbf{DeepRC hyperparameter search space.}
        Every $5\cdot10^{3}$ updates, the current model was evaluated against the validation fold.
        The early stopping hyperparameter was determined by selecting the model with the best loss
        on the validation fold after $10^{5}$ updates.
        $^*$: Experiments for $\{64;128;256\}$ kernels were omitted for datasets with  motif implantation probabilities $\geq 1\%$ in the category ``simulated immunosequencing data''.
        }%
        \label{tab:deeprc_settings}%
    \end{center}%
\end{table}

\paragraph{Known motif.}
This method does not have hyperparameters and has been applied 
to all datasets except for the \emph{CMV dataset}. 

\paragraph{SVM.}
The corresponding hyperparameter $C$ of the SVM is optimized by randomly 
drawing $10^{3}$ values in the range of $[-6; 6]$ 
according to a uniform distribution. These values act 
as the exponents of a power of $10$ and are applied 
for each of the two kernel types (see Table~\ref{tab:svm_settings}).

\begin{table}[htp]%
    \begin{center}%
        \begin{tabular}{lc}%
            \toprule
            $C$ & $10^{\{-6;6\}}$ \\
            type of kernel & $\{\mathrm{MinMax}; \mathrm{Jaccard}\}$ \\
            number of trials & $10^{3}$ \\
            \bottomrule
        \end{tabular}%
        \subcaption[Hyperparameter search of the SVM baseline]{}{Settings used in the hyperparameter search of the SVM baseline approach. 
        The number of trials defines the quantity of random values of the 
        $C$ penalty term (per type of kernel).}%
        \label{tab:svm_settings}%
    \end{center}%
\end{table}

\paragraph{KNN.}
The amount of neighbors is treated as the hyperparameter and optimized by grid search operating in the discrete range of $[1; \max\{N, 10^{3}\}]$ with a step size of $1$. The corresponding tight upper bound is automatically defined by the total amount of samples $N\in\mathbb{N}_{> 0}$ in the training set, capped at $10^{3}$ (see Table~\ref{tab:knn_settings}).

\begin{table}[htp]%
    \begin{center}%
        \begin{tabular}{lc}%
            \toprule
            number of neighbors & $\{1; \max\{N, 10^{3}\}\}$ \\
            type of kernel & $\{\mathrm{MinMax}; \mathrm{Jaccard}\}$ \\
            \bottomrule
        \end{tabular}%
        \caption[Hyperparameter search of the KNN baseline]{Settings used in the hyperparameter search of the KNN baseline approach. 
        The number of trials (per type of kernel) is automatically defined by the total 
        amount of samples $N\in\mathbb{N}_{> 0}$ in the training set, capped at $10^{3}$.}%
        \label{tab:knn_settings}%
    \end{center}%
\end{table}

\paragraph{Logistic regression.}
The hyperparameter optimization strategy that was used was grid search
across hyperparameters given in Table~\ref{tab:logistic_regression_settings}.

\begin{table}[htp]%
    \begin{center}%
        \begin{tabular}{lc}%
            \toprule
            learning rate & $10^{-\{2; 3; 4\}}$ \\
            batch size & $4$ \\
            max. updates & $10^{5}$ \\
            coefficient $\beta_{1}$ (Adam) & $0.9$ \\
            coefficient $\beta_{2}$ (Adam) & $0.999$ \\
            weight decay weightings & $\{(l1=10^{-7}, l2=10^{-3});(l1=10^{-7}, l2=10^{-5})\}$ \\
            \bottomrule
        \end{tabular}%
        \caption[Hyperparameter search of the logistic regression]{Settings used in the hyperparameter search of the logistic regression baseline approach.}%
        \label{tab:logistic_regression_settings}%
    \end{center}%
\end{table}

\paragraph{Burden test.}
The burden test selects two hyperparameters: the number of features
in the burden set and the type of features, see Table~\ref{tab:burden_settings}.

\begin{table}[htp]%
    \begin{center}%
        \begin{tabular}{lc}%
            \toprule
            number of features in burden set & $\{50,100,150,250\}$ \\
            type of features & $\{\mathrm{4MER}; \mathrm{sequence}\}$ \\
            \bottomrule
        \end{tabular}%
        \caption[Hyperparameter search of the burden test]{Settings used in the hyperparameter search of the burden test approach.}%
        \label{tab:burden_settings}%
    \end{center}%
\end{table}

\paragraph{Logistic MIL.}
For this method, we adjusted the learning rate as well as the batch size as 
hyperparameters by randomly drawing $25$ different hyperparameter 
combinations from a uniform distribution. The corresponding 
range of the learning rate is $[-4.5; -1.5]$, which
acts as the exponent of a power of $10$. The batch size lies 
within the range of $[1; 32]$. For each hyperparameter 
combination, a model is optimized by gradient descent using
Adam, whereas the early stopping parameter is adjusted 
according to the corresponding validation set (see Table~\ref{tab:mil_settings}).

\begin{table}[htp]%
    \begin{center}%
        \begin{tabular}{lc}%
            \toprule
            learning rate & $10^{\{-4.5;-1.5\}}$ \\
            batch size & $\{1;32\}$ \\
            relative abundance term & $\{\mathrm{4MER}; \mathrm{TCR}\text{\textbeta}\}$ \\
            number of trials & $25$ \\
            max. epochs & $10^{2}$ \\
            coefficient $\beta_{1}$ (Adam) & $0.9$ \\
            coefficient $\beta_{2}$ (Adam) & $0.999$ \\
            \bottomrule
        \end{tabular}%
        \caption[Hyperparameter search of the logistic MIL baseline]{Settings used in the hyperparameter search of the logistic MIL baseline approach. 
        The number of trials (per type of relative abundance) defines the quantity of combinations of random values of the learning rate 
        as well as the batch size.}%
        \label{tab:mil_settings}%
    \end{center}%
\end{table}

\clearpage
\section{Results}
\label{sec:detailed_results}
In this section, we report the detailed results on all four categories 
of datasets (a) simulated immunosequencing data (Table~\ref{tab:results_naive})
(b) LSTM-generated data (Table~\ref{tab:results_lstm}),
(c) real-world data with implanted signals (Table~\ref{tab:results_milena}), 
and (d) real-world data on the \emph{CMV dataset} (Table~\ref{tab:results_cmv}),
as discussed in the main paper.

\begingroup
\setlength{\tabcolsep}{2pt} %
\renewcommand{\arraystretch}{1.5} %
\begingroup
\setlength{\tabcolsep}{2pt} %
\renewcommand{\arraystretch}{1.5} %
\begin{table*}[h]%
    \begin{center}%
        \resizebox{0.98\textwidth}{!}{%
        \begin{tabular}{l acacacacacacacacaca}%
            \toprule
            ID &0 &1 & 2 & 3 & 4 & 5 & 6 & 7 & 8 & 9 &10 &11 &12 &13 &14 &15 &16 &17 & avg.\\
motif freq. $\rho$ & 1\% & 0.1\% & 0.01\% & 1\% & 0.1\% & 0.01\% & 1\% & 0.1\% & 0.01\% & 1\% & 0.1\% & 0.01\% & 1\% & 0.1\% & 0.01\% & 1\% & 0.1\% & 0.01\% & -- \\
implanted motif & \texttt{SFEN} & \texttt{SFEN} & \texttt{SFEN} & \texttt{SF\textsuperscript{d}EN} & \texttt{SF\textsuperscript{d}EN} & \texttt{SF\textsuperscript{d}EN} & \texttt{SFZN} & \texttt{SFZN} & \texttt{SFZN} & \texttt{SF\textsuperscript{d}ZN} & \texttt{SF\textsuperscript{d}ZN} & \texttt{SF\textsuperscript{d}ZN} & \texttt{SZZN} & \texttt{SZZN} & \texttt{SZZN} & \texttt{SZ\textsuperscript{d}ZN} & \texttt{SZ\textsuperscript{d}ZN} & \texttt{SZ\textsuperscript{d}ZN} & --\\
\midrule
DeepRC & {\bf 1.000} & {\bf 1.000} & 0.703 & {\bf 1.000} & {\bf 1.000} & 0.600 & {\bf 1.000} & {\bf 1.000} & 0.509 & {\bf 1.000} & {\bf 1.000} & 0.492 & {\bf 1.000} & {\bf 0.997} & 0.487 & 0.999 & {\bf 0.942} & 0.492 & {\bf 0.864} \\
& \footnotesize{$\pm$ 0.000} & \footnotesize{$\pm$ 0.000} & \footnotesize{$\pm$ 0.271} & \footnotesize{$\pm$ 0.000} & \footnotesize{$\pm$ 0.000} & \footnotesize{$\pm$ 0.218} & \footnotesize{$\pm$ 0.000} & \footnotesize{$\pm$ 0.000} & \footnotesize{$\pm$ 0.029} & \footnotesize{$\pm$ 0.000} & \footnotesize{$\pm$ 0.001} & \footnotesize{$\pm$ 0.017} & \footnotesize{$\pm$ 0.001} & \footnotesize{$\pm$ 0.002} & \footnotesize{$\pm$ 0.023} & \footnotesize{$\pm$ 0.001} & \footnotesize{$\pm$ 0.048} & \footnotesize{$\pm$ 0.013} & \footnotesize{$\pm$ 0.223} \\
SVM (MinMax) & {\bf 1.000} & {\bf 1.000} & 0.764 & {\bf 1.000} & {\bf 1.000} & 0.603 & {\bf 1.000} & 0.998 & {\bf 0.539} & {\bf 1.000} & 0.994 & {\bf 0.529} & {\bf 1.000} & 0.741 & {\bf 0.513} & {\bf 1.000} & 0.706 & 0.503 & 0.827 \\
& \footnotesize{$\pm$ 0.000} & \footnotesize{$\pm$ 0.000} & \footnotesize{$\pm$ 0.016} & \footnotesize{$\pm$ 0.000} & \footnotesize{$\pm$ 0.000} & \footnotesize{$\pm$ 0.021} & \footnotesize{$\pm$ 0.000} & \footnotesize{$\pm$ 0.002} & \footnotesize{$\pm$ 0.024} & \footnotesize{$\pm$ 0.000} & \footnotesize{$\pm$ 0.004} & \footnotesize{$\pm$ 0.016} & \footnotesize{$\pm$ 0.000} & \footnotesize{$\pm$ 0.024} & \footnotesize{$\pm$ 0.006} & \footnotesize{$\pm$ 0.000} & \footnotesize{$\pm$ 0.013} & \footnotesize{$\pm$ 0.013} & \footnotesize{$\pm$ 0.210} \\
SVM (Jaccard) & 0.783 & 0.505 & 0.500 & 0.656 & 0.504 & 0.492 & 0.629 & 0.499 & 0.505 & 0.594 & 0.508 & 0.497 & 0.620 & 0.496 & 0.506 & 0.595 & 0.507 & 0.505 & 0.550 \\
& \footnotesize{$\pm$ 0.010} & \footnotesize{$\pm$ 0.009} & \footnotesize{$\pm$ 0.010} & \footnotesize{$\pm$ 0.009} & \footnotesize{$\pm$ 0.018} & \footnotesize{$\pm$ 0.018} & \footnotesize{$\pm$ 0.011} & \footnotesize{$\pm$ 0.010} & \footnotesize{$\pm$ 0.009} & \footnotesize{$\pm$ 0.007} & \footnotesize{$\pm$ 0.017} & \footnotesize{$\pm$ 0.013} & \footnotesize{$\pm$ 0.007} & \footnotesize{$\pm$ 0.006} & \footnotesize{$\pm$ 0.019} & \footnotesize{$\pm$ 0.013} & \footnotesize{$\pm$ 0.012} & \footnotesize{$\pm$ 0.017} & \footnotesize{$\pm$ 0.080} \\
KNN (MinMax) & 0.669 & 0.802 & 0.503 & 0.722 & 0.757 & 0.493 & 0.766 & 0.678 & 0.496 & 0.762 & 0.652 & 0.489 & 0.797 & 0.512 & 0.498 & 0.796 & 0.511 & 0.503 & 0.634 \\
& \footnotesize{$\pm$ 0.204} & \footnotesize{$\pm$ 0.265} & \footnotesize{$\pm$ 0.038} & \footnotesize{$\pm$ 0.214} & \footnotesize{$\pm$ 0.255} & \footnotesize{$\pm$ 0.017} & \footnotesize{$\pm$ 0.241} & \footnotesize{$\pm$ 0.165} & \footnotesize{$\pm$ 0.014} & \footnotesize{$\pm$ 0.237} & \footnotesize{$\pm$ 0.139} & \footnotesize{$\pm$ 0.015} & \footnotesize{$\pm$ 0.271} & \footnotesize{$\pm$ 0.023} & \footnotesize{$\pm$ 0.014} & \footnotesize{$\pm$ 0.270} & \footnotesize{$\pm$ 0.037} & \footnotesize{$\pm$ 0.006} & \footnotesize{$\pm$ 0.129} \\
KNN (Jaccard) & 0.516 & 0.493 & 0.497 & 0.506 & 0.500 & 0.492 & 0.509 & 0.493 & 0.497 & 0.495 & 0.504 & 0.500 & 0.502 & 0.497 & 0.500 & 0.502 & 0.503 & {\bf 0.513} & 0.501 \\
& \footnotesize{$\pm$ 0.035} & \footnotesize{$\pm$ 0.020} & \footnotesize{$\pm$ 0.013} & \footnotesize{$\pm$ 0.015} & \footnotesize{$\pm$ 0.019} & \footnotesize{$\pm$ 0.014} & \footnotesize{$\pm$ 0.017} & \footnotesize{$\pm$ 0.011} & \footnotesize{$\pm$ 0.018} & \footnotesize{$\pm$ 0.013} & \footnotesize{$\pm$ 0.004} & \footnotesize{$\pm$ 0.017} & \footnotesize{$\pm$ 0.011} & \footnotesize{$\pm$ 0.017} & \footnotesize{$\pm$ 0.022} & \footnotesize{$\pm$ 0.015} & \footnotesize{$\pm$ 0.020} & \footnotesize{$\pm$ 0.012} & \footnotesize{$\pm$ 0.007} \\
Logistic regression & {\bf 1.000} & {\bf 1.000} & {\bf 0.786} & {\bf 1.000} & {\bf 1.000} & {\bf 0.607} & {\bf 1.000} & 0.997 & 0.527 & {\bf 1.000} & 0.992 & 0.526 & {\bf 1.000} & 0.719 & 0.505 & {\bf 1.000} & 0.694 & 0.510 & 0.826 \\
& \footnotesize{$\pm$ 0.000} & \footnotesize{$\pm$ 0.000} & \footnotesize{$\pm$ 0.009} & \footnotesize{$\pm$ 0.000} & \footnotesize{$\pm$ 0.000} & \footnotesize{$\pm$ 0.025} & \footnotesize{$\pm$ 0.000} & \footnotesize{$\pm$ 0.002} & \footnotesize{$\pm$ 0.018} & \footnotesize{$\pm$ 0.000} & \footnotesize{$\pm$ 0.004} & \footnotesize{$\pm$ 0.019} & \footnotesize{$\pm$ 0.000} & \footnotesize{$\pm$ 0.019} & \footnotesize{$\pm$ 0.015} & \footnotesize{$\pm$ 0.001} & \footnotesize{$\pm$ 0.021} & \footnotesize{$\pm$ 0.017} & \footnotesize{$\pm$ 0.211} \\
Logistic MIL (KMER) & {\bf 1.000} & {\bf 1.000} & 0.509 & {\bf 1.000} & 0.783 & 0.489 & {\bf 1.000} & 0.544 & 0.517 & {\bf 1.000} & 0.529 & 0.483 & 0.579 & 0.498 & 0.502 & 0.550 & 0.488 & 0.498 & 0.665 \\
& \footnotesize{$\pm$ 0.000} & \footnotesize{$\pm$ 0.000} & \footnotesize{$\pm$ 0.039} & \footnotesize{$\pm$ 0.000} & \footnotesize{$\pm$ 0.216} & \footnotesize{$\pm$ 0.023} & \footnotesize{$\pm$ 0.000} & \footnotesize{$\pm$ 0.038} & \footnotesize{$\pm$ 0.018} & \footnotesize{$\pm$ 0.000} & \footnotesize{$\pm$ 0.043} & \footnotesize{$\pm$ 0.007} & \footnotesize{$\pm$ 0.042} & \footnotesize{$\pm$ 0.017} & \footnotesize{$\pm$ 0.018} & \footnotesize{$\pm$ 0.051} & \footnotesize{$\pm$ 0.009} & \footnotesize{$\pm$ 0.005} & \footnotesize{$\pm$ 0.224} \\
Logistic MIL (TCR\textbeta) & 0.544 & 0.505 & 0.493 & 0.487 & 0.476 & 0.500 & 0.520 & 0.495 & 0.510 & 0.492 & 0.506 & 0.503 & 0.509 & 0.505 & 0.500 & 0.475 & 0.489 & 0.500 & 0.501 \\
& \footnotesize{$\pm$ 0.078} & \footnotesize{$\pm$ 0.014} & \footnotesize{$\pm$ 0.018} & \footnotesize{$\pm$ 0.021} & \footnotesize{$\pm$ 0.019} & \footnotesize{$\pm$ 0.022} & \footnotesize{$\pm$ 0.053} & \footnotesize{$\pm$ 0.009} & \footnotesize{$\pm$ 0.022} & \footnotesize{$\pm$ 0.014} & \footnotesize{$\pm$ 0.019} & \footnotesize{$\pm$ 0.010} & \footnotesize{$\pm$ 0.034} & \footnotesize{$\pm$ 0.009} & \footnotesize{$\pm$ 0.011} & \footnotesize{$\pm$ 0.013} & \footnotesize{$\pm$ 0.024} & \footnotesize{$\pm$ 0.019} & \footnotesize{$\pm$ 0.016} \\
Burden test & 0.770 & 0.523 & 0.510 & 0.666 & 0.510 & 0.509 & 0.652 & 0.508 & 0.505 & 0.583 & 0.508 & 0.509 & 0.564 & 0.508 & 0.507 & 0.536 & 0.508 & 0.504 & 0.549\\
& \footnotesize{$\pm$ 0.013} & \footnotesize{$\pm$ 0.013} & \footnotesize{$\pm$ 0.014} & \footnotesize{$\pm$ 0.011} & \footnotesize{$\pm$ 0.009} & \footnotesize{$\pm$ 0.007} & \footnotesize{$\pm$ 0.008} & \footnotesize{$\pm$ 0.011} & \footnotesize{$\pm$ 0.012} & \footnotesize{$\pm$ 0.012} & \footnotesize{$\pm$ 0.007} & \footnotesize{$\pm$ 0.014} & \footnotesize{$\pm$ 0.017} & \footnotesize{$\pm$ 0.010} & \footnotesize{$\pm$ 0.020} & \footnotesize{$\pm$ 0.012} & \footnotesize{$\pm$ 0.016} & \footnotesize{$\pm$ 0.016} & \footnotesize{$\pm$ 0.074} \\
\midrule
Known motif b. & 1.000 & 1.000 & 0.973 & 1.000 & 1.000 & 0.865 & 1.000 & 1.000 & 0.700 & 1.000 & 0.989 & 0.609 & 1.000 & 0.946 & 0.570 & 1.000 & 0.834 & 0.532 & 0.890 \\
& \footnotesize{$\pm$ 0.000} & \footnotesize{$\pm$ 0.000} & \footnotesize{$\pm$ 0.004} & \footnotesize{$\pm$ 0.000} & \footnotesize{$\pm$ 0.000} & \footnotesize{$\pm$ 0.004} & \footnotesize{$\pm$ 0.000} & \footnotesize{$\pm$ 0.000} & \footnotesize{$\pm$ 0.020} & \footnotesize{$\pm$ 0.000} & \footnotesize{$\pm$ 0.002} & \footnotesize{$\pm$ 0.017} & \footnotesize{$\pm$ 0.000} & \footnotesize{$\pm$ 0.010} & \footnotesize{$\pm$ 0.024} & \footnotesize{$\pm$ 0.000} & \footnotesize{$\pm$ 0.016} & \footnotesize{$\pm$ 0.020} & \footnotesize{$\pm$ 0.168} \\
Known motif c. & 0.999 & 0.720 & 0.529 & 0.999 & 0.698 & 0.534 & 0.999 & 0.694 & 0.532 & 1.000 & 0.696 & 0.527 & 0.997 & 0.666 & 0.520 & 0.998 & 0.668 & 0.509 & 0.738 \\
& \footnotesize{$\pm$ 0.001} & \footnotesize{$\pm$ 0.014} & \footnotesize{$\pm$ 0.020} & \footnotesize{$\pm$ 0.001} & \footnotesize{$\pm$ 0.013} & \footnotesize{$\pm$ 0.017} & \footnotesize{$\pm$ 0.001} & \footnotesize{$\pm$ 0.012} & \footnotesize{$\pm$ 0.012} & \footnotesize{$\pm$ 0.001} & \footnotesize{$\pm$ 0.018} & \footnotesize{$\pm$ 0.018} & \footnotesize{$\pm$ 0.002} & \footnotesize{$\pm$ 0.010} & \footnotesize{$\pm$ 0.009} & \footnotesize{$\pm$ 0.002} & \footnotesize{$\pm$ 0.012} & \footnotesize{$\pm$ 0.013} & \footnotesize{$\pm$ 0.202} \\

      \bottomrule
        \end{tabular}%
        }%
        \caption[AUC estimates for all 18 datasets in "simulated immunosequencing data"]{AUC estimates based on 5-fold CV for all $18$ datasets in category ``simulated immunosequencing data''.
        The reported errors are standard deviations across the $5$ cross-validation folds except 
        for the last column ``avg.'', in which they show standard deviations across datasets.
        Wildcard characters in motifs are indicated by \texttt{Z},
        characters with $50\%$ probability of being removed by \texttt{\textsuperscript{d}.}
        }%
        \label{tab:results_naive}%
    \end{center}%
\end{table*}
\endgroup

\clearpage

\begin{table}[htp]
    \centering
    \resizebox{0.98\textwidth}{!}{%
    \begin{tabular}{lacacac}
    \toprule
ID & 0 & 1 & 2 & 3 & 4 & avg.  \\
motif freq. $\rho$ & 10\% & 1\% & 0.5\% & 0.1\% & 0.05\% & -- \\
implanted motif & \texttt{G\textsuperscript{r}S\textsuperscript{r}A\textsuperscript{r}F\textsuperscript{r}} & \texttt{G\textsuperscript{r}S\textsuperscript{r}A\textsuperscript{r}F\textsuperscript{r}} &\texttt{G\textsuperscript{r}S\textsuperscript{r}A\textsuperscript{r}F\textsuperscript{r}} &\texttt{G\textsuperscript{r}S\textsuperscript{r}A\textsuperscript{r}F\textsuperscript{r}} &\texttt{G\textsuperscript{r}S\textsuperscript{r}A\textsuperscript{r}F\textsuperscript{r}} & --\\
\midrule
DeepRC & {\bf 1.000} \footnotesize{$\pm$ 0.000} & {\bf 1.000} \footnotesize{$\pm$ 0.000} & {\bf 1.000} \footnotesize{$\pm$ 0.000} & {\bf 1.000} \footnotesize{$\pm$ 0.000} & {\bf 0.998} \footnotesize{$\pm$ 0.002} & {\bf 1.000} \footnotesize{$\pm$ 0.001} \\
SVM (MinMax) & {\bf 1.000} \footnotesize{$\pm$ 0.000} & {\bf 1.000} \footnotesize{$\pm$ 0.000} & 0.999 \footnotesize{$\pm$ 0.001} & 0.999 \footnotesize{$\pm$ 0.002} & 0.985 \footnotesize{$\pm$ 0.014} & 0.997 \footnotesize{$\pm$ 0.007} \\
SVM (Jaccard) & 0.981 \footnotesize{$\pm$ 0.041} & {\bf 1.000} \footnotesize{$\pm$ 0.000} & {\bf 1.000} \footnotesize{$\pm$ 0.000} & 0.904 \footnotesize{$\pm$ 0.036} & 0.768 \footnotesize{$\pm$ 0.068} & 0.931 \footnotesize{$\pm$ 0.099} \\
KNN (MinMax) & 0.699 \footnotesize{$\pm$ 0.272} & 0.717 \footnotesize{$\pm$ 0.263} & 0.732 \footnotesize{$\pm$ 0.263} & 0.536 \footnotesize{$\pm$ 0.156} & 0.516 \footnotesize{$\pm$ 0.153} & 0.640 \footnotesize{$\pm$ 0.105} \\
KNN (Jaccard) & 0.698 \footnotesize{$\pm$ 0.285} & 0.606 \footnotesize{$\pm$ 0.237} & 0.523 \footnotesize{$\pm$ 0.164} & 0.550 \footnotesize{$\pm$ 0.186} & 0.539 \footnotesize{$\pm$ 0.194} & 0.583 \footnotesize{$\pm$ 0.071} \\
Logistic regression & {\bf 1.000} \footnotesize{$\pm$ 0.000} & {\bf 1.000} \footnotesize{$\pm$ 0.000} & 0.934 \footnotesize{$\pm$ 0.147} & 0.604 \footnotesize{$\pm$ 0.193} & 0.427 \footnotesize{$\pm$ 0.156} & 0.793 \footnotesize{$\pm$ 0.262} \\
Logistic MIL (KMER) & 0.997 \footnotesize{$\pm$ 0.004} & 0.718 \footnotesize{$\pm$ 0.112} & 0.637 \footnotesize{$\pm$ 0.144} & 0.571 \footnotesize{$\pm$ 0.146} & 0.528 \footnotesize{$\pm$ 0.129} & 0.690 \footnotesize{$\pm$ 0.186} \\
Logistic MIL (TCR\textbeta) & 0.541 \footnotesize{$\pm$ 0.086} & 0.566 \footnotesize{$\pm$ 0.162} & 0.468 \footnotesize{$\pm$ 0.086} & 0.505 \footnotesize{$\pm$ 0.067} & 0.500 \footnotesize{$\pm$ 0.121} & 0.516 \footnotesize{$\pm$ 0.038} \\
Burden test & {\bf 1.000} \footnotesize{$\pm$ 0.000} & {\bf 1.000} \footnotesize{$\pm$ 0.000} & {\bf 1.000} \footnotesize{$\pm$ 0.000} & 0.999 \footnotesize{$\pm$ 0.003} & 0.792 \footnotesize{$\pm$ 0.280} & 0.958 \footnotesize{$\pm$ 0.093} \\
\midrule
Known motif b. & 1.000 \footnotesize{$\pm$ 0.000} & 1.000 \footnotesize{$\pm$ 0.000} & 1.000 \footnotesize{$\pm$ 0.000} & 0.999 \footnotesize{$\pm$ 0.003} & 0.999 \footnotesize{$\pm$ 0.003} & 1.000 \footnotesize{$\pm$ 0.001} \\
Known motif c. & 1.000 \footnotesize{$\pm$ 0.000} & 1.000 \footnotesize{$\pm$ 0.000} & 0.989 \footnotesize{$\pm$ 0.011} & 0.722 \footnotesize{$\pm$ 0.085} & 0.626 \footnotesize{$\pm$ 0.094} & 0.867 \footnotesize{$\pm$ 0.180} \\ 
\bottomrule
        \end{tabular} 
        }
    \caption[AUC estimates for all 5 datasets in ``LSTM-generated data'']{
            AUC estimates based on 5-fold CV for all $5$ datasets in category ``LSTM-generated data''.
            The reported errors are standard deviations across the $5$ cross-validation folds except 
            for the last column ``avg.'', in which they show standard deviations across datasets.
            Characters affected by noise, as described in \ref{sec:datasets}, paragraph ``LSTM-generated data'', are indicated by \texttt{\textsuperscript{r}.}
            }
    \label{tab:results_lstm}
\end{table}

\begin{table*}[htp]
    \centering
    \begin{tabular}{lacaca}
    \toprule
  & OM 1\% & OM 0.1\%  & MM 1\% & MM 0.1\% & Avg.  \\
  \midrule
DeepRC & {\bf 1.000} \footnotesize{$\pm$ 0.000} & {\bf 0.984} \footnotesize{$\pm$ 0.008} & 0.999 \footnotesize{$\pm$ 0.001} & {\bf 0.938} \footnotesize{$\pm$ 0.009} & {\bf 0.980} \footnotesize{$\pm$ 0.029} \\
SVM (MinMax) & {\bf 1.000} \footnotesize{$\pm$ 0.000} & 0.578 \footnotesize{$\pm$ 0.020} & {\bf 1.000} \footnotesize{$\pm$ 0.000} & 0.531 \footnotesize{$\pm$ 0.019} & 0.777 \footnotesize{$\pm$ 0.258} \\
SVM (Jaccard) & 0.988 \footnotesize{$\pm$ 0.003} & 0.527 \footnotesize{$\pm$ 0.016} & {\bf 1.000} \footnotesize{$\pm$ 0.000} & 0.574 \footnotesize{$\pm$ 0.019} & 0.772 \footnotesize{$\pm$ 0.257} \\
KNN (MinMax) & 0.744 \footnotesize{$\pm$ 0.237} & 0.486 \footnotesize{$\pm$ 0.031} & 0.674 \footnotesize{$\pm$ 0.182} & 0.500 \footnotesize{$\pm$ 0.022} & 0.601 \footnotesize{$\pm$ 0.128} \\
KNN (Jaccard) & 0.652 \footnotesize{$\pm$ 0.155} & 0.484 \footnotesize{$\pm$ 0.025} & 0.695 \footnotesize{$\pm$ 0.200} & 0.508 \footnotesize{$\pm$ 0.025} & 0.585 \footnotesize{$\pm$ 0.104} \\
Logistic regression & {\bf 1.000} \footnotesize{$\pm$ 0.000} & 0.544 \footnotesize{$\pm$ 0.035} & 0.991 \footnotesize{$\pm$ 0.003} & 0.512 \footnotesize{$\pm$ 0.035} & 0.762 \footnotesize{$\pm$ 0.270} \\
Logistic MIL (KMER) & 0.541 \footnotesize{$\pm$ 0.074} & 0.506 \footnotesize{$\pm$ 0.034} & 0.994 \footnotesize{$\pm$ 0.004} & 0.620 \footnotesize{$\pm$ 0.153} & 0.665 \footnotesize{$\pm$ 0.224} \\
Logistic MIL (TCR\textbeta) & 0.503 \footnotesize{$\pm$ 0.032} & 0.501 \footnotesize{$\pm$ 0.016} & 0.992 \footnotesize{$\pm$ 0.003} & 0.782 \footnotesize{$\pm$ 0.030} & 0.695 \footnotesize{$\pm$ 0.238} \\
Burden test & {\bf 1.000} \footnotesize{$\pm$ 0.000} & 0.640 \footnotesize{$\pm$ 0.048} & {\bf 1.000} \footnotesize{$\pm$ 0.000} & 0.891 \footnotesize{$\pm$ 0.016} & 0.883 \footnotesize{$\pm$ 0.170} \\
\midrule
Known motif b. & 1.000 \footnotesize{$\pm$ 0.000} & 0.704 \footnotesize{$\pm$ 0.028} & 0.994 \footnotesize{$\pm$ 0.003} & 0.620 \footnotesize{$\pm$ 0.038} & 0.830 \footnotesize{$\pm$ 0.196} \\
Known motif c. & 0.920 \footnotesize{$\pm$ 0.004} & 0.562 \footnotesize{$\pm$ 0.028} & 0.647 \footnotesize{$\pm$ 0.030} & 0.515 \footnotesize{$\pm$ 0.031} & 0.661 \footnotesize{$\pm$ 0.181} \\ 
   \bottomrule
    \end{tabular}
    \caption[AUC estimates for all $4$ datasets in ``real-world data with implanted signals'']{
            AUC estimates based on 5-fold CV for all $4$ datasets in category ``real-world data with implanted signals''.
            The reported errors are standard deviations across the $5$ cross-validation folds except 
            for the last column ``avg.'', in which they show standard deviations across datasets.
            \textbf{OM 1\%:} In this dataset, a single motif with a frequency of 1\% was implanted. 
            \textbf{OM 0.1\%:} In this dataset, a single motif with a frequency of 0.1\% was implanted.
            \textbf{MM 1\%:} In this dataset, multiple motifs with a frequency of 1\% were implanted.
            \textbf{MM 0.1\%:} In this dataset, multiple motifs with a frequency of 0.1\% were implanted.
            A detailed description of the motifs is provided in Section~\ref{sec:datasets}, paragraph ``Real-world data with implanted signals.''.}
    \label{tab:results_milena}
\end{table*}

\begin{table}[htp]
    \centering
    \begin{tabular}{lacac}
    \toprule
    & AUC & F1 score & Balanced accuracy & Accuracy \\
    \midrule
    DeepRC & {\bf0.831} \footnotesize{$\pm$ 0.002} & {\bf 0.726} \footnotesize{$\pm$ 0.050} & {\bf 0.741} \footnotesize{$\pm$ 0.043} & 0.727 \footnotesize{$\pm$ 0.049} \\
    SVM (MinMax) & 0.825 \footnotesize{$\pm$ 0.022} & 0.680 \footnotesize{$\pm$ 0.056} & 0.734 \footnotesize{$\pm$ 0.037} & {\bf 0.742} \footnotesize{$\pm$ 0.031} \\
    SVM (Jaccard) & 0.546 \footnotesize{$\pm$ 0.021} & 0.272 \footnotesize{$\pm$ 0.184} & 0.523 \footnotesize{$\pm$ 0.026} & 0.542 \footnotesize{$\pm$ 0.032} \\
    KNN (MinMax) & 0.679 \footnotesize{$\pm$ 0.076} & 0.000 \footnotesize{$\pm$ 0.000} & 0.500 \footnotesize{$\pm$ 0.000} & 0.545 \footnotesize{$\pm$ 0.044} \\
    KNN (Jaccard) & 0.534 \footnotesize{$\pm$ 0.039} & 0.073 \footnotesize{$\pm$ 0.101} & 0.508 \footnotesize{$\pm$ 0.012} & 0.551 \footnotesize{$\pm$ 0.042} \\
    Logistic regression & 0.607 \footnotesize{$\pm$ 0.058} & 0.244 \footnotesize{$\pm$ 0.206} & 0.552 \footnotesize{$\pm$ 0.049} & 0.590 \footnotesize{$\pm$ 0.019} \\
    Logistic MIL (KMER) & 0.582 \footnotesize{$\pm$ 0.065} & 0.118 \footnotesize{$\pm$ 0.264} & 0.503 \footnotesize{$\pm$ 0.007} & 0.515 \footnotesize{$\pm$ 0.058} \\
    Logistic MIL (TCR\textbeta) & 0.515 \footnotesize{$\pm$ 0.073} & 0.000 \footnotesize{$\pm$ 0.000} & 0.496 \footnotesize{$\pm$ 0.008} & 0.541 \footnotesize{$\pm$ 0.039} \\
    Burden test & 0.699 \footnotesize{$\pm$ 0.041} & - & - & - \\
    \bottomrule
    \end{tabular} 
    \caption[Results on the CMV dataset given by AUC, F1 score, balanced accuracy, and accuracy]{Results on the \emph{CMV dataset} (real-world data) in terms of AUC, F1 score, balanced accuracy, and accuracy. For F1 score, balanced accuracy, and accuracy, all methods use their default thresholds. Each entry shows mean and standard deviation across $5$ 
    cross-validation folds.}
    \label{tab:results_cmv}
\end{table}

\clearpage
\section{Repertoire generation via LSTM}
\label{appsec:lstm}
We trained a conventional next-character LSTM model \citep{graves2013generating} based on the implementation in \url{https://github.com/spro/practical-pytorch} (access date 1st of May, 2020) using \texttt{PyTorch 1.3.1} \citep{paszke2019pytorch}.
For this, we applied an LSTM model with $100$ LSTM blocks in $2$ layers,
which was trained for $5,000$ epochs using the Adam optimizer \citep{kingma2014adam}
with learning rate $0.01$,
an input batch size of $100$ character chunks,
and a character chunk length of $200$.
As input we used the immuno-sequences in the \texttt{CDR3} column of the \emph{CMV dataset},
where we repeated sequences according to their counts in the repertoires,
as specified in the \texttt{templates} column of the \emph{CMV dataset}.
We excluded repertoires with unknown CMV status and unknown sequence abundance from training.

After training,
we generated $1,000$ repertoires using a \texttt{temperature} value of $0.8$.
The number of sequences per repertoire was sampled from 
a Gaussian $\mathcal N(\mu=285k, \sigma=156k)$ distribution,
where the whole repertoire was generated by the LSTM at once.
That is,
the LSTM can base the generation of the individual AA sequences in a repertoire,
including the AAs and the lengths of the sequences,
on the generated repertoire.
A random immuno-sequence from the trained-on repertoires was used as initialization for the generation process.
This immuno-sequence was not included in the generated repertoire.

Finally, we randomly assigned $500$ of the generated repertoires to the positive (diseased) and $500$ to the negative (healthy) class.
We then implanted motifs in the positive class repertoires
as described in Section~\ref{subsec:lstm_generated_data}.

As illustrated in the comparison of histograms given in Fig.~\ref{tab:lstm_gen_hists},
the generated immuno-sequences exhibit a very similar distribution of 4-mers and AAs
compared to the original \emph{CMV dataset}.

\begin{figure}[htp]
    \begin{center}
    \begingroup
    \setlength{\tabcolsep}{4pt} %
    \renewcommand{\arraystretch}{1.5} %
    \resizebox{1\textwidth}{!}{%
    \begin{tabular}{ll}
    \multicolumn{1}{c}{\textbf{Real-world data}} & \multicolumn{1}{c}{\textbf{LSTM-generated data}} \\
    \multicolumn{1}{c}{\textbf{a)}} & \multicolumn{1}{c}{\textbf{b)}}\\
    \includegraphics[width=0.45\textwidth]{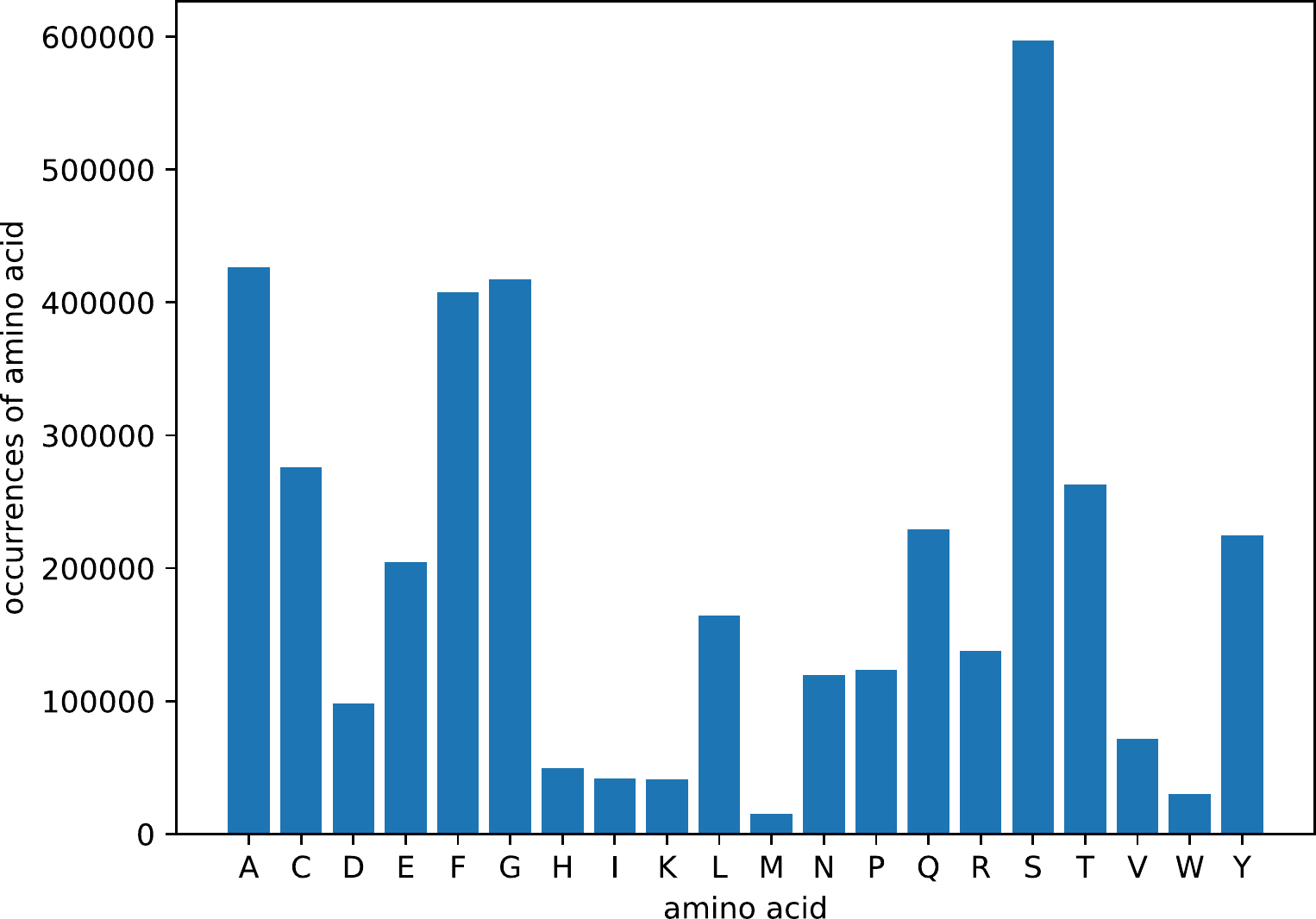}& \includegraphics[width=0.45\textwidth]{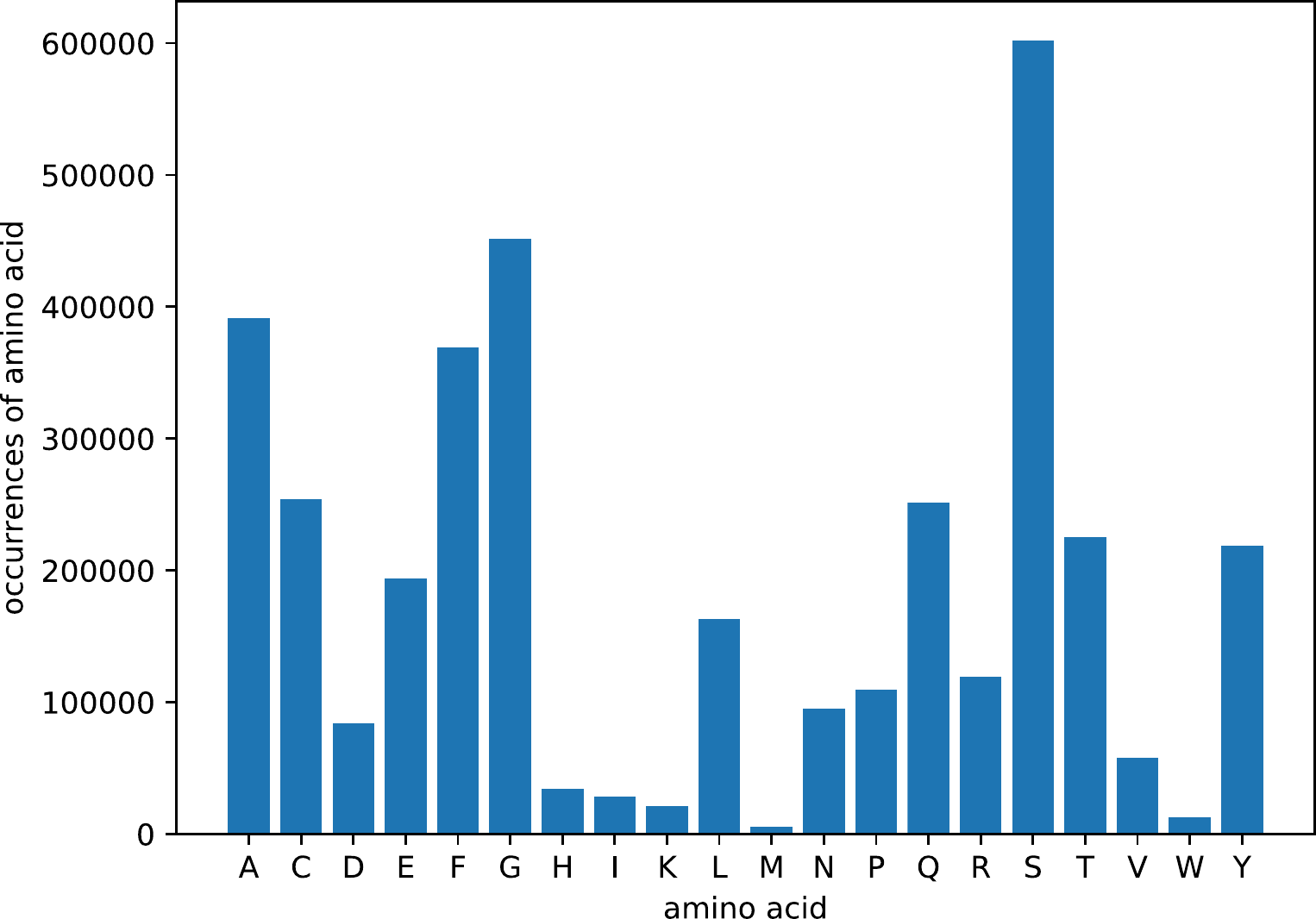}\\
    \multicolumn{1}{c}{\textbf{c)}} & \multicolumn{1}{c}{\textbf{d)}}\\
    \includegraphics[width=0.45\textwidth]{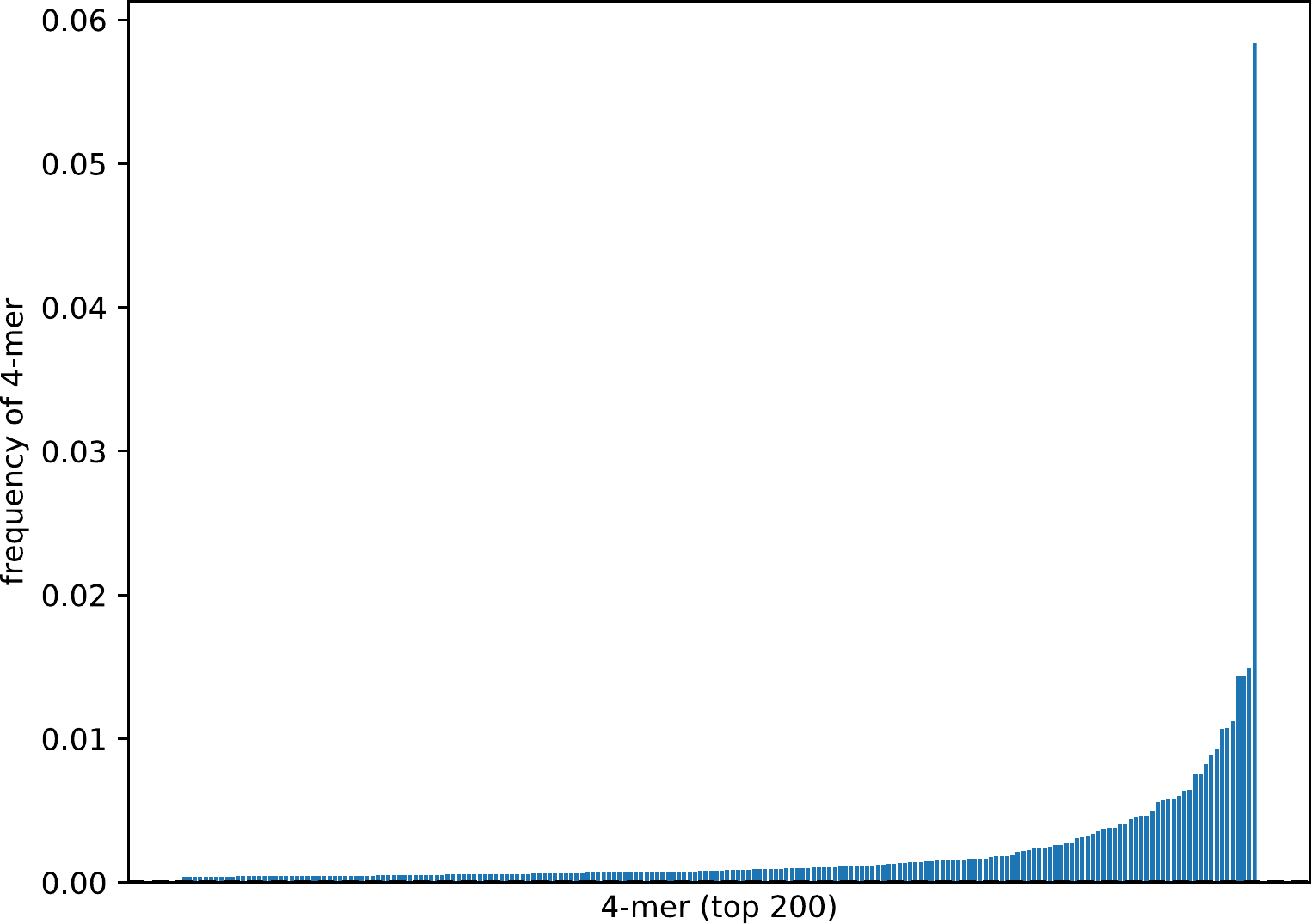}&\includegraphics[width=0.45\textwidth]{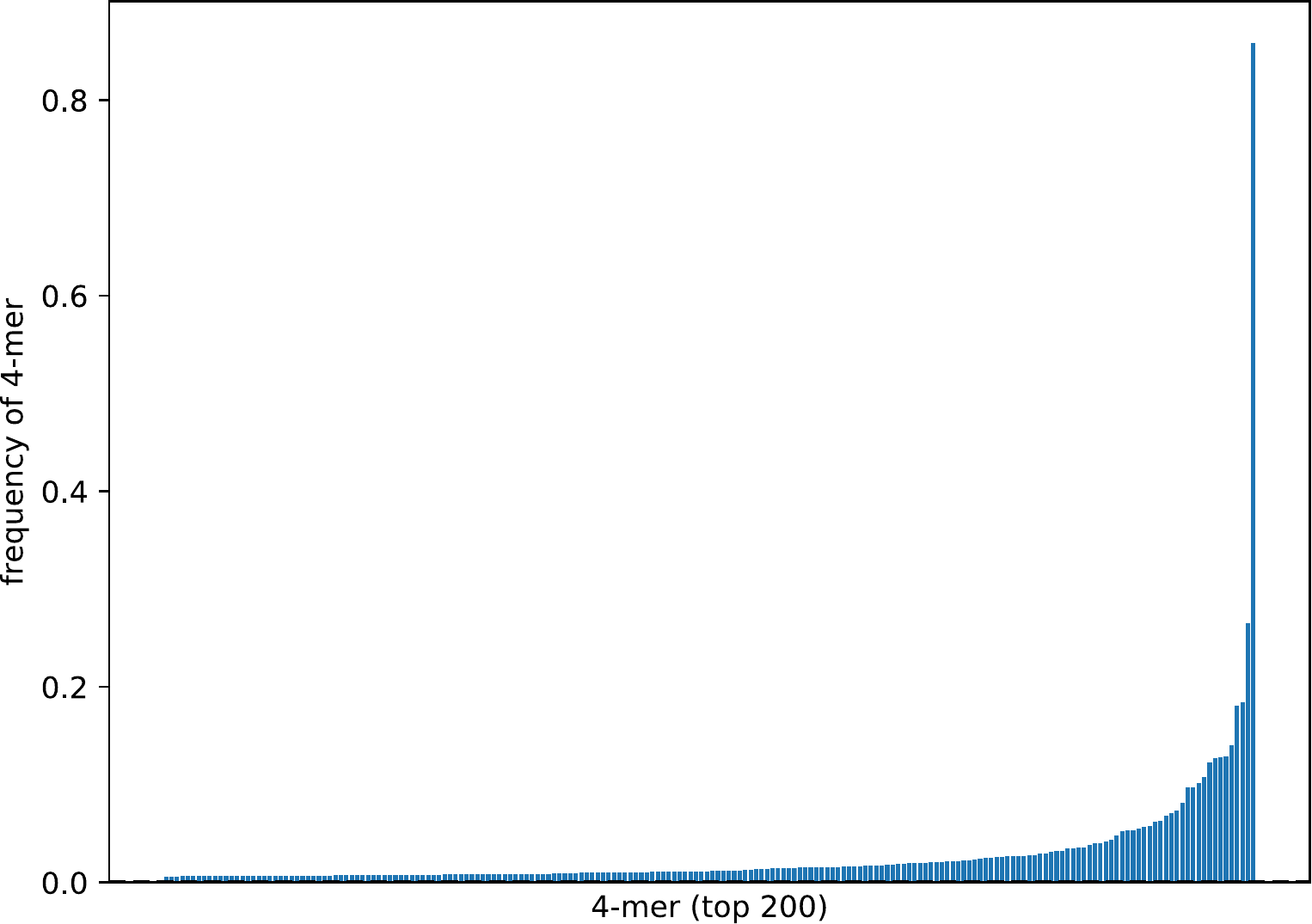}\\
    \multicolumn{1}{c}{\textbf{e)}} & \multicolumn{1}{c}{\textbf{f)}}\\
    \includegraphics[width=0.45\textwidth]{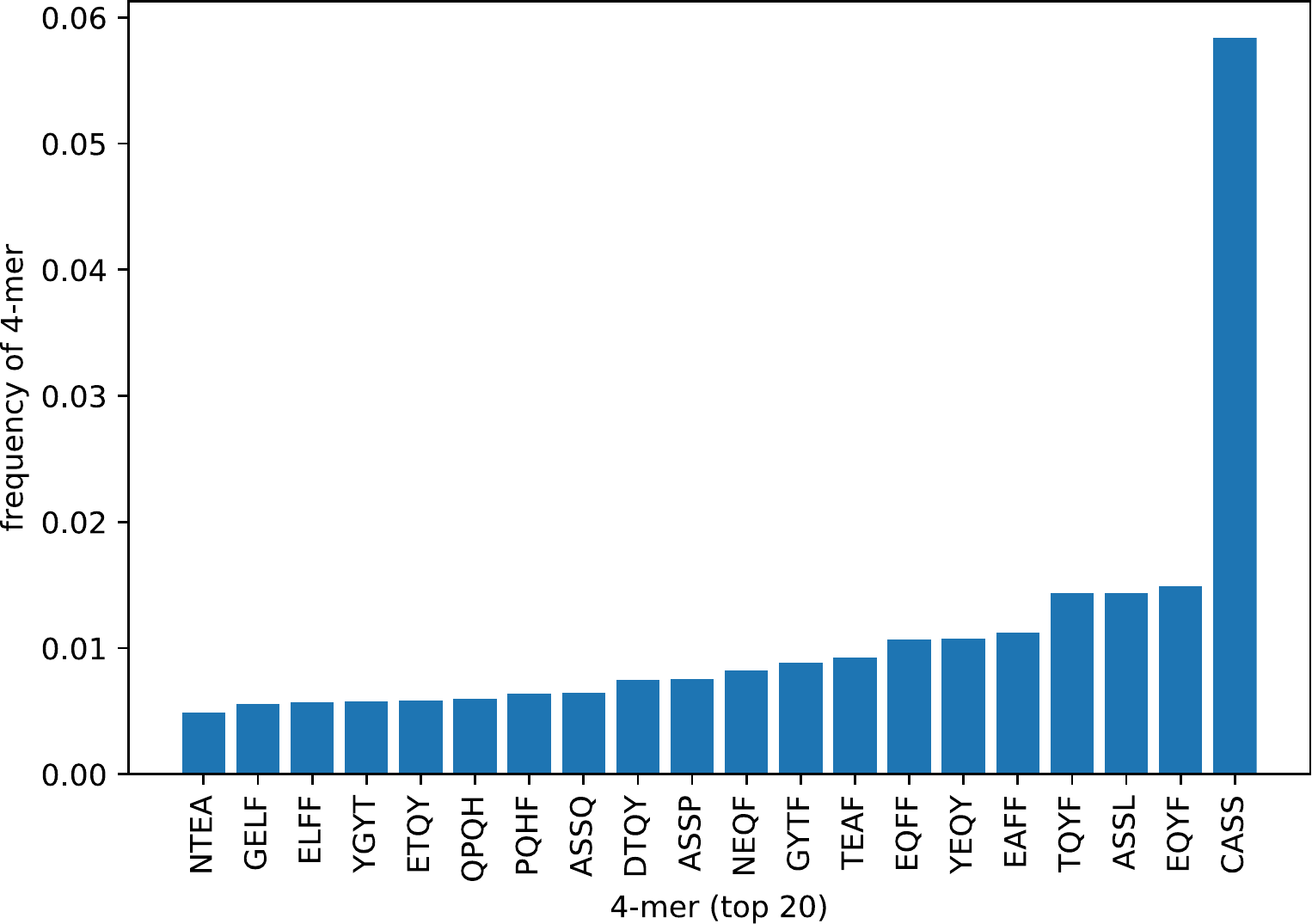}&
    \includegraphics[width=0.45\textwidth]{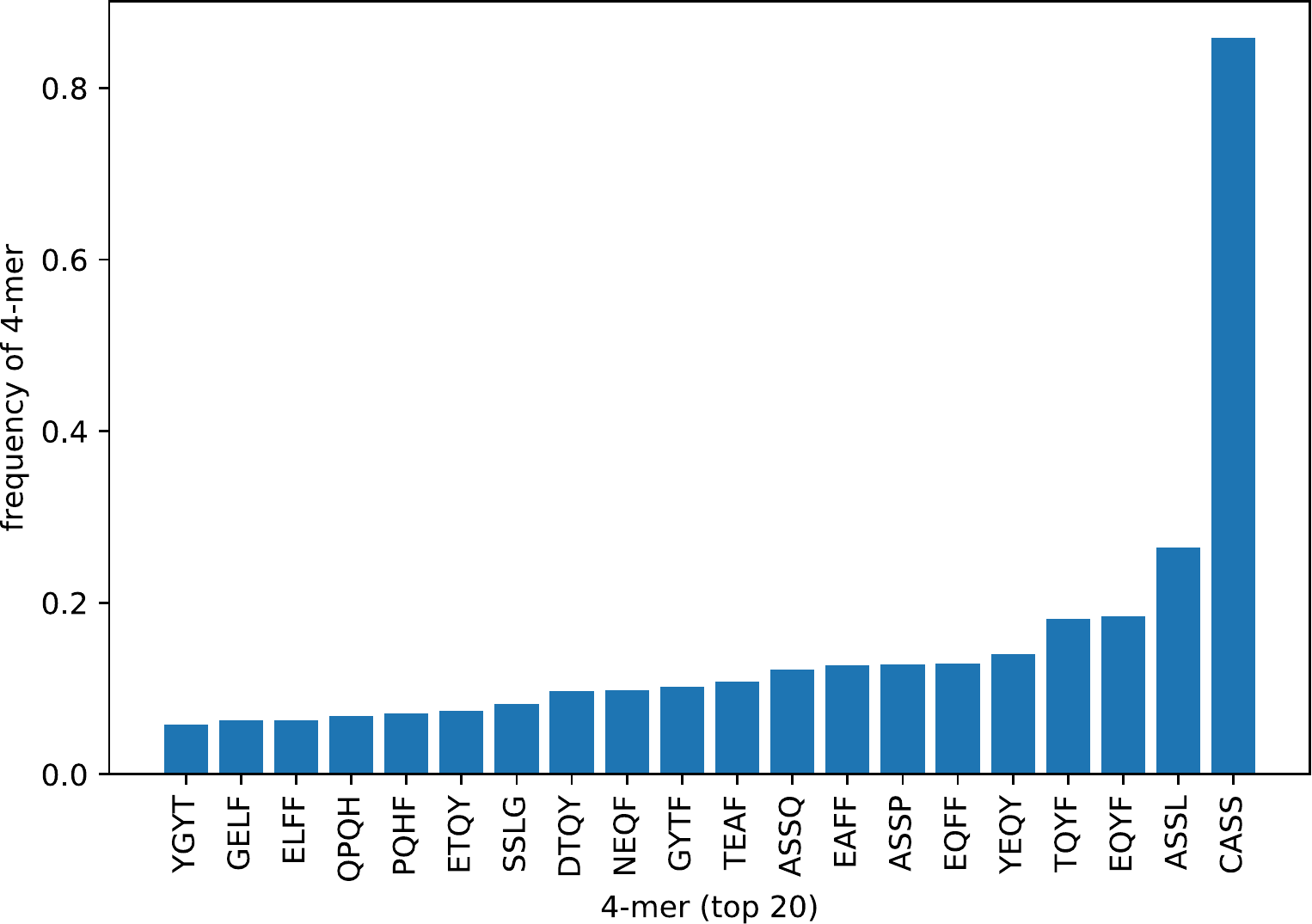}\\
    \end{tabular}
    }%
    \endgroup
   \caption[Distribution of AAs and k-mers]{Distribution of AAs and k-mers in 
   real-world \emph{CMV dataset} and 
   LSTM-generated data.
   \textbf{Left:} Histograms of real-world data.
   \textbf{Right:} Histograms of LSTM-generated data.
   \textbf{a)} Frequency of AAs in sequences of the \emph{CMV dataset}.
   \textbf{b)} Frequency of AAs in sequences of the LSTM-generated datasets.
   \textbf{c)} Frequency of top 200 4-mers in sequences of the \emph{CMV dataset}.
   \textbf{d)} Frequency of top 200 4-mers in sequences of the LSTM-generated datasets.
   \textbf{e)} Frequency of top 20 4-mers in sequences of the \emph{CMV dataset}.
   \textbf{f)} Frequency of top 20 4-mers in sequences of the LSTM-generated datasets. 
   Overall the distributions of AAs and 4-mers are similar in both datasets.}
   \label{tab:lstm_gen_hists}
   \end{center}
\end{figure}

\clearpage
\section{Interpreting DeepRC}
\label{appsec:integrated_gradients}
DeepRC allows for two forms of interpretability methods.
(a) Due to its attention-based design,
a trained model can be used to compute the attention weights of a sequence, which directly indicates its importance.
(b) DeepRC furthermore allows for the usage of contribution analysis 
methods, such as Integrated Gradients (IG) 
\citep{sundararajan2017axiomatic} or Layer-Wise Relevance 
Propagation \citep{montavon2018methods,arras2019explaining,montavon2019layer,preuer2019interpretable}.
We apply IG to identify the input patterns that are 
relevant for the classification.
To identify AA patterns with high contributions in the input sequences,
we apply IG to the AAs in the input sequences.
Additionally, we apply IG to the kernels of the 1D CNN,
which allows us to identify AA motifs with high contributions.
In detail, 
we compute the IG contributions for 
the AAs and positional features in the kernels for every repertoire in the validation and test set,
so as to exclude potential artifacts caused by  over-fitting.
Averaging the IG values over these repertoires then results in concise AA motifs.
We include qualitative visual analyses of the IG method 
on different datasets below.

Here, we provide examples for the interpretation of trained DeepRC models
using Integrated Gradients (IG) \citep{sundararajan2017axiomatic} as contribution analysis method.
The following illustrations were created using $50$ IG steps,
which we found sufficient to achieve stable IG results.

A visual analysis of DeepRC models on the simulated datasets,
as illustrated in Tab.~\ref{tab:simulations_kernel_ig} and Fig.~\ref{fig:simulations_input_ig},
shows that the implanted motifs can be successfully extracted from the trained model
and are straight-forward to interpret.
In the real-world \emph{CMV dataset},
DeepRC finds complex patterns with high variability in the center regions of the immuno-sequences,
as illustrated in figure~\ref{fig:ig_cmv_inputs}.

\begin{table}[htp]%
    \begin{center}%
        \begingroup
        \setlength{\tabcolsep}{10pt} %
        \renewcommand{\arraystretch}{1.5} %
        \resizebox{0.98\textwidth}{!}{%
        \begin{tabular}{lcccc}
            \toprule
            & \multicolumn{4}{c}{\textbf{Simulated}}  \\
            extracted motif %
            & \includegraphics[angle=0,width=0.1\textwidth]{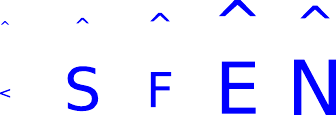}%
            & \includegraphics[angle=0,width=0.1\textwidth]{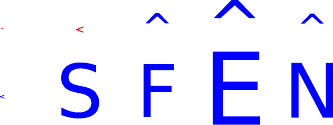}%
            & \includegraphics[angle=0,width=0.1\textwidth]{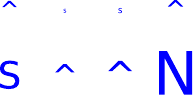}
            & \includegraphics[angle=0,width=0.1\textwidth]{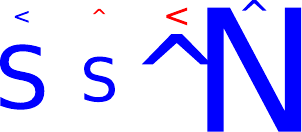}\\
            implanted motif(s) & \texttt{SFEN} & \texttt{SF\textsuperscript{d}EN} & \texttt{SZZN} & \texttt{SZ\textsuperscript{d}ZN} \\
            motif freq. $\rho$ & $0.01\%$ & $0.01\%$ & $0.1\%$ & $0.1\%$
            \\
            \\
            & \multicolumn{1}{c}{\textbf{LSTM-generated}} & & \multicolumn{2}{c}{\textbf{Real-world data with implanted signals}} \\
            extracted motif %
            & \includegraphics[angle=0,width=0.1\textwidth]{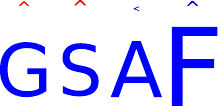}%
            & & \includegraphics[angle=0,width=0.1\textwidth]{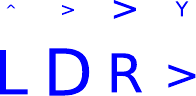}%
            & \includegraphics[angle=0,width=0.2\textwidth]{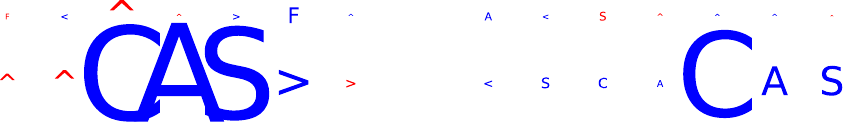}\\
            implanted motif(s) & \texttt{G\textsuperscript{r}S\textsuperscript{r}A\textsuperscript{r}F\textsuperscript{r}}%
            & & \texttt{L\textsuperscript{r}D\textsuperscript{r}R\textsuperscript{r}} & \{\texttt{L\textsuperscript{r}D\textsuperscript{r}R\textsuperscript{r}}; \texttt{C\textsuperscript{r}A\textsuperscript{r}S}; \texttt{G\textsuperscript{r}L-N}\} \\
            motif freq. $\rho$ & $0.05\%$ & & $0.1\%$ & $0.1\%$ \\
            \bottomrule
        \end{tabular}
                    }%
        \endgroup
    \end{center}%
    \caption[Visualization of extracted motifs]{Visualization of motifs extracted from trained DeepRC models for datasets from categories
    ``simulated immunosequencing data'',
    ``LSTM-generated data'', and
    ``real-world data with implanted signals''.
    Motif extraction was performed using Integrated Gradients on the 1D CNN kernels over the validation set and test set repertoires of one CV fold.
    Wildcard characters are indicated by \texttt{Z},
    random noise on characters by \texttt{\textsuperscript{r}},
    characters with $50\%$ probability of being removed by \texttt{\textsuperscript{d}}, and
    gap locations of random lengths of $\{0;1;2\}$ by \texttt{-}.
    Larger characters in the extracted motifs indicate higher contribution,
    with blue indicating positive contribution and red indicating negative contribution towards the prediction of the diseased class.
    Contributions to positional encoding are indicated by \texttt{$<$} (beginning of sequence), \texttt{$\wedge$} (center of sequence), and \texttt{$>$} (end of sequence).
    Only kernels with relatively high contributions are shown,
    i.e. with contributions roughly greater than the average contribution of all kernels.
    \label{tab:simulations_kernel_ig}}%
\end{table}

\begin{figure}[htp]
    \begin{center}
        \begingroup
        
        \setlength{\tabcolsep}{10pt} %
        \renewcommand{\arraystretch}{1.5} %
        \begin{tabular}{ll}
    \textbf{a)} & \includegraphics[height=16pt]{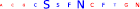}\\
    \textbf{b)} & \includegraphics[height=16pt]{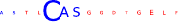}\\
    \textbf{c)} & \includegraphics[height=16pt]{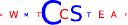}
        \end{tabular}
        
        \endgroup
   \caption[Interpretation of the DeepRC classifier]{Integrated Gradients applied to input sequences of positive class repertoires.
   Three sequences with the highest contributions to the prediction of their respective repertoires are shown.
   \textbf{a)} Input sequence taken from ``simulated immunosequencing data''
   with implanted motif \texttt{SZ\textsuperscript{d}Z\textsuperscript{d}N} 
   and motif implantation probability $0.1\%$.
   The DeepRC model reacts to the \texttt{S} and \texttt{N} at the 5\textsuperscript{th} and 8\textsuperscript{th} sequence position,
   thereby identifying the implanted motif in this sequence. 
   \textbf{b)} and \textbf{c)} Input sequence taken from ``real-world data with implanted signals''
   with implanted motifs \{\texttt{L\textsuperscript{r}D\textsuperscript{r}R\textsuperscript{r}}; \texttt{C\textsuperscript{r}A\textsuperscript{r}S}; \texttt{G\textsuperscript{r}L-N}\} 
   and motif implantation probability $0.1\%$.
   The DeepRC model reacts to the fully implanted motif \texttt{CAS} (b)
   and to the partly implanted motif AAs \texttt{C} and \texttt{A} at the 5\textsuperscript{th} and 7\textsuperscript{th} sequence position (c),
   thereby identifying the implanted motif in the sequences.
    Wildcard characters in implanted motifs are indicated by \texttt{Z},
    characters with $50\%$ probability of being removed by \texttt{\textsuperscript{d}}, and
    gap locations of random lengths of $\{0;1;2\}$ by \texttt{-}.
    Larger characters in the sequences indicate higher contribution,
    with blue indicating positive contribution and red indicating negative contribution towards the prediction of the diseased class.
   \label{fig:simulations_input_ig}}
   \end{center}
\end{figure}

\begin{figure}[htp]
    \begin{center}
    \includegraphics[height=0.8\textheight]{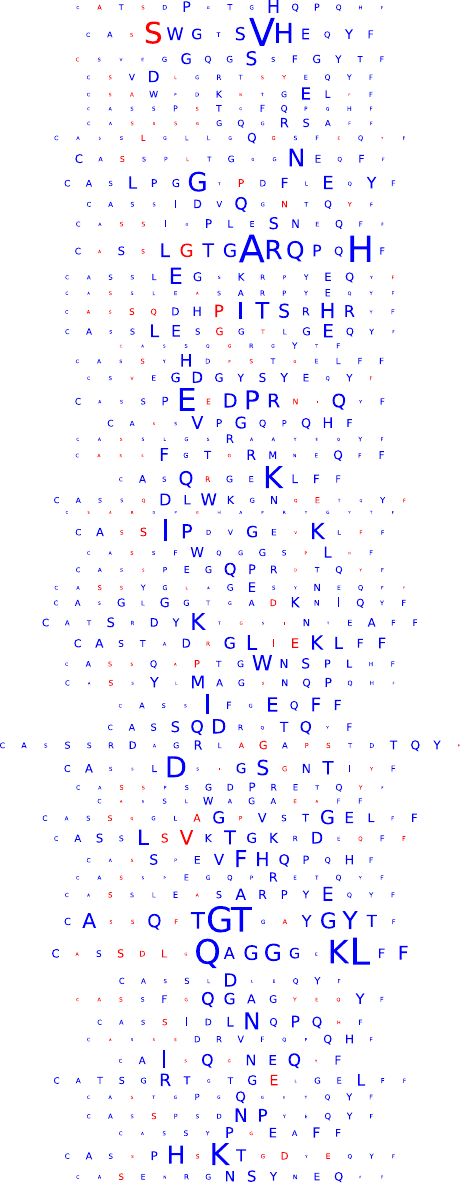}
   \caption[Visualization of the contributions of AA]{
   Visualization of the contributions of characters within a sequence via IG.
   Each sequence was selected from a different repertoire and 
   showed the highest contribution in its repertoire.
   The model was trained on \emph{CMV dataset},
   using a kernel size of $9$, $32$ kernels
   and $137$ repertoires for early stopping.
    Larger characters in the extracted motifs indicate higher contribution,
    with blue indicating positive contribution and red indicating negative contribution towards the prediction of the disease class.
   \label{fig:ig_cmv_inputs}}
   \end{center}
\end{figure}

\clearpage
\section{Attention values for previously associated CMV sequences}
\label{sec:attention_cmv}
Table~\ref{tab:emerson_recovered} lists sequences of the \emph{CMV
dataset} that were previously associated with CMV immune status
and their assigned high attention values by DeepRC.

\begin{table}[h]
    \centering
    \resizebox{0.99\textwidth}{!}{%
    \begin{tabular}{clccabaaclccabaa}
    \toprule
index & sequence & attention & quantile & index & sequence & attention & quantile & index & sequence & attention & quantile & index & sequence & attention & quantile \\
\midrule
1 & CASSGQGAYEQYF & 1.000 & 0.999 & 42 & CASSLGGAGDTQYF & 1.000 & 1.000 & 83 & CASSYVRTGGNYGYTF & 0.967 & 0.932 & 124 & CASSLTGGNSGNTIYF & 0.991 & 0.977 \\
2 & CASSIGPLEHNEQFF & 0.947 & 0.900 & 43 & CASNRDRGRYEQYF & 0.991 & 0.978 & 84 & CASSLAGVDYEQYF & 0.999 & 0.996 & 125 & CASSRNRGQETQYF & 0.978 & 0.952 \\
3 & CASSPDRVGQETQYF & 0.995 & 0.987 & 44 & CSVRDNHNQPQHF & 0.965 & 0.929 & 85 & CASSLGAGNQPQHF & 1.000 & 0.999 & 126 & CASSLGQGLAEAFF & 0.996 & 0.989 \\
4 & CASSLEAEYEQYF & 0.992 & 0.980 & 45 & CASSAQGAYEQYF & 0.998 & 0.995 & 86 & CASSRDRNYGYTF & 0.998 & 0.995 & 127 & CASRTGESGYTF & 0.985 & 0.965 \\
5 & CASSIEGNQPQHF & 0.993 & 0.983 & 46 & CATSRGTVSYEQYF & 0.990 & 0.975 & 87 & CASGRDTYEQYF & 0.999 & 0.997 & 128 & CASSSDSGGTDTQYF & 0.951 & 0.906 \\
6 & CATSDGDEQFF & 0.998 & 0.996 & 47 & CASSPPSGLTDTQYF & 0.978 & 0.951 & 88 & CAWSVSDLAKNIQYF & 0.954 & 0.911 & 129 & CASSVDGGRGTEAFF & 0.995 & 0.987 \\
7 & CASSLVAGGRETQYF & 0.988 & 0.971 & 48 & CASSGDRLYEQYF & 0.998 & 0.994 & 89 & CASSPNQETQYF & 0.999 & 0.996 & 130 & CSVEVRGTDTQYF & 0.955 & 0.912 \\
8 & CASSRGRQETQYF & 0.997 & 0.993 & 49 & CASSLNRGQETQYF & 0.996 & 0.988 & 90 & CSASDHEQYF & 0.995 & 0.986 & 131 & CASSESGDPSSYEQYF & 0.980 & 0.955 \\
9 & CASSAGQGVTYEQYF & 0.998 & 0.995 & 50 & CASSLGVGPYNEQFF & 0.986 & 0.967 & 91 & CASSWDRDNSPLHF & 0.918 & 0.855 & 132 & CASSEEAGGSGYTF & 0.982 & 0.959 \\
10 & CASSQNRGQETQYF & 0.995 & 0.987 & 51 & CATSDSVTNTGELFF & 0.989 & 0.973 & 92 & CASSPGQEAGANVLTF & 0.823 & 0.728 & 133 & CAISESQDRGHEQYF & 0.823 & 0.728 \\
11 & CASSPQRNTEAFF & 1.000 & 0.999 & 52 & CASSRNRESNQPQHF & 0.968 & 0.934 & 93 & CASSLVAAGRETQYF & 0.959 & 0.919 & 134 & CASSPTGGELFF & 0.989 & 0.974 \\
12 & CASSLAPGATNEKLFF & 0.976 & 0.949 & 53 & CASSEARTRAFF & 0.927 & 0.869 & 94 & CASSPHRNTEAFF & 0.999 & 0.998 & 135 & CASSVETGGTEAFF & 0.995 & 0.986 \\
13 & CASSLIGVSSYNEQFF & 0.983 & 0.961 & 54 & CASSYNPYSNQPQHF & 0.892 & 0.819 & 95 & CASRGQGWDEKLFF & 0.994 & 0.984 & 136 & CASASANYGYTF & 0.816 & 0.720 \\
14 & CSVRDNFNQPQHF & 0.915 & 0.851 & 55 & CASSLGHRDSSYEQYF & 0.987 & 0.969 & 96 & CASSQVETDTQYF & 0.994 & 0.984 & 137 & CASSSRTGEETQYF & 0.996 & 0.988 \\
15 & CASSQTGGRNQPQHF & 0.997 & 0.992 & 56 & CASSRLAASTDTQYF & 0.992 & 0.979 & 97 & CASRDWDYTDTQYF & 0.994 & 0.984 & 138 & CASSLGRGYEKLFF & 0.985 & 0.965 \\
16 & CASSLVIGGDTEAFF & 0.966 & 0.931 & 57 & CASSVTGGTDTQYF & 1.000 & 0.999 & 98 & CASSSDRVGQETQYF & 0.980 & 0.955 & 139 & CASSGLNEQFF & 0.994 & 0.984 \\
17 & CASSLRREKLFF & 0.998 & 0.993 & 58 & CASSPPGQGSDTQYF & 0.975 & 0.946 & 99 & CASSLGDRPDTQYF & 0.940 & 0.889 & 140 & CASSRNRAQETQYF & 0.994 & 0.984 \\
18 & CASSFHGFNQPQHF & 0.991 & 0.978 & 59 & CATSDSRTGGQETQYF & 0.900 & 0.829 & 100 & CASSLEGQGFGYTF & 0.944 & 0.895 & 141 & CASTPGDEQFF & 0.988 & 0.971 \\
19 & CATSRDTQGSYGYTF & 0.917 & 0.854 & 60 & CASSSPGRSGANVLTF & 0.995 & 0.986 & 101 & CASSSGQVYGYTF & 0.999 & 0.996 & 142 & CASSLGIDTQYF & 0.997 & 0.991 \\
20 & CASSRLAGGTDTQYF & 0.999 & 0.998 & 61 & CASSPLSDTQYF & 0.998 & 0.994 & 102 & CASSEEGIQPQHF & 0.998 & 0.994 & 143 & CASSIRTNYYGYTF & 0.996 & 0.990 \\
21 & CASSFPTSGQETQYF & 0.982 & 0.959 & 62 & CASSLTGGRNQPQHF & 0.999 & 0.997 & 103 & CASSLETYGYTF & 0.998 & 0.995 & 144 & CASSPISNEQFF & 0.967 & 0.933 \\
22 & CASSPGDEQYF & 0.998 & 0.993 & 63 & CASSIQGYSNQPQHF & 0.993 & 0.983 & 104 & CASSFPGGETQYF & 0.992 & 0.979 & 145 & CASSQNRAQETQYF & 0.984 & 0.962 \\
23 & CASSLPSGLTDTQYF & 0.994 & 0.985 & 64 & CASSTTGGDGYTF & 0.978 & 0.952 & 105 & CASSSGQVQETQYF & 0.997 & 0.993 & 146 & CASSALGGAGTGELFF & 0.985 & 0.964 \\
24 & CASSEIPNTEAFF & 0.997 & 0.992 & 65 & CASSVLAGPTDTQYF & 0.951 & 0.906 & 106 & CASSEGARQPQHF & 0.999 & 0.998 & 147 & CASSLAVLPTDTQYF & 0.996 & 0.989 \\
25 & CASSIWGLDTEAFF & 0.959 & 0.919 & 66 & CASSHRDRNYEQYF & 0.987 & 0.969 & 107 & CSALGHSNQPQHF & 0.926 & 0.867 & 148 & CASSLQAGANEQFF & 0.969 & 0.935 \\
26 & CASSPGDEQFF & 0.999 & 0.997 & 67 & CASSPSRNTEAFF & 0.999 & 0.998 & 108 & CASSLLWDQPQHF & 0.986 & 0.967 & 149 & CASSTGGAQPQHF & 0.998 & 0.993 \\
27 & CATSRDSQGSYGYTF & 0.980 & 0.955 & 68 & CASSLGGPGDTQYF & 0.993 & 0.982 & 109 & CASSLVGDGYTF & 1.000 & 1.000 & 150 & CASSLGASGSRTDTQYF & 0.932 & 0.876 \\
28 & CASSYGGLGSYEQYF & 0.995 & 0.987 & 69 & CASSEARGGVEKLFF & 0.989 & 0.974 & 110 & CASSSRGTGELFF & 0.999 & 0.997 & 151 & CASSRGTGATDTQYF & 0.999 & 0.998 \\
29 & CASSPSTGTEAFF & 0.997 & 0.992 & 70 & CASSTGTSGSYEQYF & 0.999 & 0.998 & 111 & CATSRVAGETQYF & 0.980 & 0.955 & 152 & CASSYPGETQYF & 0.997 & 0.992 \\
30 & CSVEEDEGIYGYTF & 0.964 & 0.927 & 71 & CASRSDSGANVLTF & 0.973 & 0.942 & 112 & CASRGQGAGELFF & 0.987 & 0.969 & 153 & CASSLTDTGELFF & 0.994 & 0.984 \\
31 & CASSPAGLNTEAFF & 0.996 & 0.988 & 72 & CASSLEAENEQFF & 0.973 & 0.943 & 113 & CASSPGGTQYF & 0.999 & 0.996 & 154 & CASRPQGNYGYTF & 0.998 & 0.996 \\
32 & CASSLGLKGTQYF & 0.964 & 0.928 & 73 & CASSEAPSTSTDTQYF & 0.989 & 0.973 & 114 & CASSLQGINQPQHF & 0.999 & 0.997 & 155 & CASSTSGNTIYF & 1.000 & 0.999 \\
33 & CASMGGASYEQYF & 0.991 & 0.978 & 74 & CASSLQGADTQYF & 0.997 & 0.991 & 115 & CASSQGRHTDTQYF & 0.960 & 0.921 & 156 & CASSSGTGDEQYF & 1.000 & 1.000 \\
34 & CASSQVPGQGDNEQFF & 0.983 & 0.961 & 75 & CASSLEGQQPQHF & 0.994 & 0.984 & 116 & CASSPRWQETQYF & 0.991 & 0.978 & 157 & CASSPPAGTNYGYTF & 0.947 & 0.900 \\
35 & CATSDGDTQYF & 0.996 & 0.989 & 76 & CASSYGGEGYTF & 0.999 & 0.996 & 117 & CASRDRDRVNTEAFF & 0.970 & 0.938 & 158 & CASSPLGGTTEAFF & 0.995 & 0.988 \\
36 & CATSDGETQYF & 0.998 & 0.994 & 77 & CASSLRGSSYNEQFF & 0.999 & 0.998 & 118 & CASSWDRGTEAFF & 0.999 & 0.999 & 159 & CASSLGWTEAFF & 0.999 & 0.997 \\
37 & CSVRDNYNQPQHF & 0.998 & 0.993 & 78 & CASSISAGEAFF & 0.992 & 0.979 & 119 & CASSRPGQGNTEAFF & 0.994 & 0.984 & 160 & CATSREGSGYEQYF & 0.987 & 0.969 \\
38 & CASSLVASGRETQYF & 0.997 & 0.991 & 79 & CASRPTGYEQYF & 0.987 & 0.969 & 120 & CASSPGSGANVLTF & 0.999 & 0.997 & 161 & CASSYAGDGYTF & 0.992 & 0.980 \\
39 & CSASPGQGASYGYTF & 0.987 & 0.969 & 80 & CAWRGTGNSPLHF & 0.964 & 0.927 & 121 & CASRRGSSYEQYF & 0.999 & 0.998 & 162 & CASSDRGNTGELFF & 0.995 & 0.986 \\
40 & CASSESGHRNQPQHF & 0.999 & 0.997 & 81 & CASSLGDRAYNEQFF & 0.996 & 0.988 & 122 & CASRTDSGANVLTF & 0.994 & 0.986 & 163 & CSARRGPGELFF & 0.839 & 0.749 \\
41 & CASSLGHRDPNTGELFF & 0.981 & 0.958 & 82 & CASSLQGYSNQPQHF & 1.000 & 0.999 & 123 & CASSQDPRGTEAFF & 0.950 & 0.905 & 164 & CASSQGLQETQYF & 0.996 & 0.990 \\
\bottomrule
\end{tabular} 
}
    \caption[TCR\textbeta\ sequences re-discovered by DeepRC]{
    TCR\textbeta\ sequences that had been discovered by \citet{emerson2017immunosequencing} with their associated
    attention values by DeepRC. These sequences have significantly ($p$-value 1.3e-93) higher attention values than other sequences. The column "quantile" provides the quantile values 
    of the empiricial distribution of attention values across all 
    sequences in the dataset.}
    \label{tab:emerson_recovered}
\end{table} 

\clearpage
\section{DeepRC variations and ablation study}\label{sec:deeprc_variations}
In this section we investigate the impact of different variations of DeepRC on the performance on the \emph{CMV dataset}.
We consider both a CNN-based sequence embedding, as used in the main paper, and
an LSTM-based sequence embedding.
In both cases we vary the number of attention heads and
the $\beta$ parameter for the softmax function the attention mechanism (see Eq.~2 in main paper).
For the CNN-based sequence embedding we also vary the number of CNN kernels and the kernel sizes used in the 1D CNN.
For the LSTM-based sequence embedding we use one one-directional LSTM layer,
of which the output values at the last sequence position (without padding) are taken as embedding of the sequence.
Here we vary the number of LSTM blocks in the LSTM layer.
To counter over-fitting due to the increased complexity of these DeepRC variations,
we added a $l2$ weight penalty to the training loss.
The factor with which the $l2$ weight penalty contributes to the training loss is varied over $3$ orders of magnitudes,
where suitable value ranges were manually determined on one of the training folds beforehand.

To reduce the computational effort,
we do not consider all numbers of kernels that were considered in the main paper.
Furthermore, we only compute the AUC scores on 3 of the 5 cross-validation folds.
The hyperparameters,
which were used in a grid search procedure,
are listed in Tab.~\ref{tab:deeprc_variation_settings_cnn} for the CNN-based sequence 
embedding and Tab.~\ref{tab:deeprc_variation_settings_lstm} for the LSTM-based sequence embedding.

\paragraph{Results.}
We show performance in terms of AUC score with single hyperparameters set to fixed values
so as to investigate their influence in
Tab.~\ref{tab:deeprc_variations_cnn} for the CNN-based sequence embedding 
and Tab.~\ref{tab:deeprc_variations_lstm} for the LSTM-based sequence embedding.
We note that due to restricted computational resources 
this study was conducted with fewer different numbers of CNN kernels,
with the AUC estimated from only $3$ of the $5$ cross-validation folds,
which leads to a slight decrease of performance in comparison to the full hyperparameter 
search and cross-validation procedure used in the main paper.
As can be seen in Tab.~\ref{tab:deeprc_variations_cnn} and~\ref{tab:deeprc_variations_lstm},
the LSTM-based sequence embedding generalizes slightly better than the CNN-based sequence embedding.
The performance of DeepRC, however, remains rather robust w.r.t. the different hyperparameter settings.

\begin{table}[htp]%
    \begin{center}%
        \begin{tabular}{lc}%
            \toprule
            learning rate & $10^{-4}$ \\
            number of attention heads & $\{1;16;64\}$ \\
            $\beta$ of attention softmax & $\{0.1;1.0;10.0\}$ \\
            $l2$ weight penalty & $\{1.0;0.1;0.01\}$ \\
            number of kernels & $\{8;32;128\}$ \\
            number of CNN layers & $\{1\}$ \\
            number of layers in key-NN & $\{2\}$ \\
            number of units in key-NN & $\{32\}$ \\
            kernel size & $\{5;7;9\}$ \\
            subsampled seqences & $10,000$ \\
            batch size & $4$ \\
            \bottomrule
        \end{tabular}%
        \caption[Hyperparameter search space for DeepRC variations]{Hyperparameter search space for DeepRC variations with CNN-based sequence embedding.
        Every $5\cdot10^{3}$ updates, the current model was evaluated against the validation fold.
        The early stopping hyperparameter was determined by selecting the model with the best loss
        on the validation fold after $2\cdot10^{5}$ updates.
        }%
        \label{tab:deeprc_variation_settings_cnn}%
    \end{center}%
\end{table}

\begin{table}[htp]%
    \begin{center}%
        \begin{tabular}{lc}%
            \toprule
            learning rate & $10^{-4}$ \\
            number of attention heads & $\{1;16;64\}$ \\
            $\beta$ of attention softmax & $\{0.1;1.0;10.0\}$ \\
            $l2$ weight penalty & $\{0.01;0.001;0.0001\}$ \\
            number of LSTM blocks  & $\{8;32;128\}$ \\
            number of CNN layers & $\{1\}$ \\
            number of layers in key-NN & $\{2\}$ \\
            number of units in key-NN & $\{32\}$ \\
            subsampled seqences & $10,000$ \\
            batch size & $4$ \\
            \bottomrule
        \end{tabular}%
        \caption[Hyperparameter search space for DeepRC variations with LSTM embedding]{Hyperparameter search space for DeepRC variations with LSTM-based sequence embedding.
        Every $5\cdot10^{3}$ updates, the current model was evaluated against the validation fold.
        The early stopping hyperparameter was determined by selecting the model with the best loss
        on the validation fold after $2\cdot10^{5}$ updates.
        }%
        \label{tab:deeprc_variation_settings_lstm}%
    \end{center}%
\end{table}

\begin{table}[htp]%
    \begin{center}%
        \begin{tabular}{laaacccaaa}%
            \toprule
            \textbf{Fixed parameter}&\multicolumn{3}{a}{\textbf{Test set}}&\multicolumn{3}{c}{\textbf{Validation set}}&\multicolumn{3}{a}{\textbf{Training set}}\\
            &mean&&std&mean&&std&mean&&std\\
            \midrule
            beta=0.1&0.827&$\pm$&0.02&0.846&$\pm$&0.033&0.976&$\pm$&0.015\\
            beta=1.0&0.82&$\pm$&0.012&0.853&$\pm$&0.031&0.979&$\pm$&0.016\\
            beta=10.0&0.823&$\pm$&0.014&0.858&$\pm$&0.033&0.934&$\pm$&0.026\\
            \midrule
            heads=1&0.838&$\pm$&0.033&0.856&$\pm$&0.029&0.966&$\pm$&0.012\\
            heads=16&0.817&$\pm$&0.015&0.853&$\pm$&0.028&0.972&$\pm$&0.026\\
            heads=64&0.823&$\pm$&0.014&0.858&$\pm$&0.033&0.934&$\pm$&0.026\\
            \midrule
            lstms=8&0.818&$\pm$&0.011&0.837&$\pm$&0.025&0.881&$\pm$&0.013\\
            lstms=32&0.814&$\pm$&0.015&0.853&$\pm$&0.029&0.948&$\pm$&0.033\\
            lstms=128&0.818&$\pm$&0.018&0.859&$\pm$&0.032&0.943&$\pm$&0.028\\
            \bottomrule
        \end{tabular}%
        \caption[Impact of hyperparameters on DeepRC with LSTM]{Impact of hyperparameters on DeepRC with LSTM for sequence encoding.
        Mean (``mean'') and standard deviation (``std'') for the area under the ROC curve over the first $3$ folds of a 5-fold nested cross-validation for different sub-sets of hyperparameters (``sub-set'') are shown.
        The following sub-sets were considered:
        ``full'': Full grid search over hyperparameters;
        ``beta=*'': Grid search over hyperparameters with reduction to specific value $*$ of beta value of attention softmax;
        ``heads=*'': Grid search over hyperparameters with reduction to specific number $*$ of attention heads;
        ``lstms=*'': Grid search over hyperparameters with reduction to specific number $*$ of LSTM blocks for sequence embedding.
        }%
        \label{tab:deeprc_variations_lstm}%
    \end{center}%
\end{table}

\begin{table}[htp]%
    \begin{center}%
        \begin{tabular}{laaacccaaa}%
            \toprule
            \textbf{Fixed parameter}&\multicolumn{3}{a}{\textbf{Test set}}&\multicolumn{3}{c}{\textbf{Validation set}}&\multicolumn{3}{a}{\textbf{Training set}}\\
            &mean&&std&mean&&std&mean&&std\\
            \midrule
            beta=0.1&0.833&$\pm$&0.031&0.86&$\pm$&0.025&0.94&$\pm$&0.018\\
            beta=1.0&0.799&$\pm$&0.007&0.873&$\pm$&0.017&0.954&$\pm$&0.005\\
            beta=10.0&0.817&$\pm$&0.02&0.87&$\pm$&0.022&0.962&$\pm$&0.034\\
            \midrule
            heads=1&0.822&$\pm$&0.036&0.869&$\pm$&0.022&0.943&$\pm$&0.032\\
            heads=16&0.808&$\pm$&0.01&0.871&$\pm$&0.025&0.965&$\pm$&0.019\\
            heads=64&0.796&$\pm$&0.039&0.864&$\pm$&0.018&0.927&$\pm$&0.024\\
            \midrule
            ksize=5&0.822&$\pm$&0.036&0.866&$\pm$&0.021&0.926&$\pm$&0.026\\
            ksize=7&0.817&$\pm$&0.02&0.87&$\pm$&0.022&0.962&$\pm$&0.034\\
            ksize=9&0.821&$\pm$&0.016&0.869&$\pm$&0.025&0.95&$\pm$&0.031\\
            \midrule
            kernels=8&0.825&$\pm$&0.024&0.86&$\pm$&0.027&0.928&$\pm$&0.019\\
            kernels=32&0.801&$\pm$&0.001&0.877&$\pm$&0.018&0.974&$\pm$&0.017\\
            kernels=128&0.824&$\pm$&0.027&0.864&$\pm$&0.023&0.931&$\pm$&0.062\\
            \bottomrule
        \end{tabular}%
        \caption[Impact of hyperparameters on DeepRC with 1D CNN]{Impact of hyperparameters on DeepRC with 1D CNN for sequence encoding.
        Mean (``mean'') and standard deviation (``std'') for the area under the ROC curve over the first $3$ folds of a 5-fold nested cross-validation for different sub-sets of hyperparameters (``sub-set'') are shown.
        The following sub-sets were considered:
        ``full'': Full grid search over hyperparameters;
        ``beta=*'': Grid search over hyperparameters with reduction to specific value $*$ of beta value of attention softmax;
        ``heads=*'': Grid search over hyperparameters with reduction to specific number $*$ of attention heads;
        ``ksize=*'': Grid search over hyperparameters with reduction to specific kernel size $*$ of 1D CNN kernels for sequence embedding;
        ``kernels=*'': Grid search over hyperparameters with reduction to specific number $*$ of 1D CNN kernels for sequence embedding.
        }%
        \label{tab:deeprc_variations_cnn}%
    \end{center}%
\end{table}

\clearpage
\bibliographystyle{icml2020}
\bibliography{joint_bib}

\end{document}